\documentclass{article}

\usepackage{PRIMEarxiv}

\usepackage{times}
\usepackage{microtype}
\usepackage{graphicx}
\usepackage{caption}
\usepackage{subcaption}
\usepackage{booktabs} %
\usepackage{amsmath,amsfonts,amsthm,bm}       %
\usepackage{stmaryrd} %
\usepackage{xcolor}
\usepackage{algorithm}
\usepackage{algpseudocode}

\usepackage{hyperref}

\usepackage{diagbox}

\usepackage{wrapfig}
\usepackage{rotating}

\usepackage{amsmath}
\usepackage{mathtools}
\usepackage{amsthm}
\usepackage{xifthen}

\usepackage{natbib}

\usepackage[capitalize,noabbrev]{cleveref}

\usepackage[utf8]{inputenc} %
\usepackage[T1]{fontenc}    %
\usepackage{hyperref}       %
\usepackage{url}            %
\usepackage{booktabs}       %
\usepackage{amsfonts}       %
\usepackage{nicefrac}       %
\usepackage{microtype}      %
\usepackage{lipsum}
\usepackage{fancyhdr}       %

\usepackage{preamble}

\graphicspath{{figures/}}     %

\title{Improving the interpretability of GNN predictions through conformal-based graph sparsification}

\author{
	Pablo~S\'anchez-Mart\'in \\
	Max Planck Institute for Intelligent Systems \& \\
	Saarland University \\
	Tübingen/Saarbrücken, Germany\\
	\texttt{psanchez@tue.mpg.de} \\
	\And
	Kinaan Aamir Khan \\
	McGill University\\
	Montreal, Canada\\
	\texttt{kinaan.khan@mail.mcgill.ca} \\
	\And
	Isabel Valera \\
	Saarland University \\
	Max Planck Institute for Software Systems \\
	Saarbrücken, Germany\\
	\texttt{ivalera@cs.uni-saarland.de} \\
}

\begin{document}
	\maketitle

	\begin{abstract}

Graph Neural Networks (GNNs) have achieved state-of-the-art performance in solving graph classification tasks.
However, most GNN architectures aggregate information from all nodes and edges in a graph, regardless of their relevance to the task at hand, thus hindering the interpretability of their predictions.
In contrast to prior work, in this paper we propose a GNN \emph{training} approach that jointly i) finds the most predictive subgraph by removing edges and/or nodes----\emph{without making assumptions about the subgraph structure}---while ii) optimizing the performance of the graph classification task.
To that end, we rely on reinforcement learning to solve the resulting bi-level optimization with a reward function based on conformal predictions to account for the current in-training uncertainty of the classifier.
Our empirical results on nine different graph classification datasets show that our method competes in performance with baselines while relying on significantly sparser subgraphs, leading to more interpretable GNN-based predictions.

\end{abstract}

	
	\section{Introduction} 
\label{sec:intro}

Graph Neural Networks (GNNs) have become a cornerstone in modern machine learning, excelling in diverse domains such as social network analysis \cite{Cao2019PopularityPO, Cao2019CoupledGN}, recommender systems \cite{Fan2019GraphNN, Bai2020TemporalGN} and bioinformatics \cite{Guo2021FewShotGL, Ramirez2020ClassificationOC, Xiong2020PushingTB}. However, the complexity of GNNs contributes to one of their principal shortcoming: a lack of human-interpretable predictions. 
This opacity hinders their potential for practical, real-world impact, as practitioners often require interpretable models to inform decision-making, ensure trustworthiness, and comply with regulations \cite{DoshiVelez2017TowardsAR}.
Current interpretability methods for GNN predictors aim to identify the most predictive subgraph from an original graph \cite{SUGAR}. 
In essence, interpretability in this context is closely tied to graph sparsity, implying the use of a minimal set of nodes and edges from the graph for prediction, rather than the entire graph $\Gcal$.

\begin{figure}
  \centering
  \includegraphics[width=0.8\textwidth]{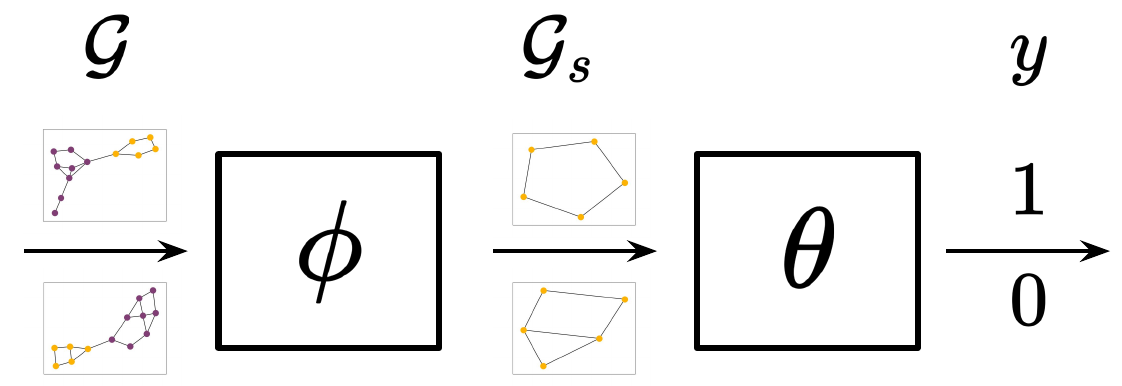}
  \captionsetup{font=small}
  \caption{\textbf{\ours\ pipeline.} Illustration of the pipeline of \ours\ using the synthetic \bashapes\, designed for binary graph classification. On the left, we present two examples of original graphs, denoted as $\Gcal$, corresponding to the positive class (top, cycle motif) and negative class (bottom, house motif). The policy $\phi$ of \ours\ takes $\Gcal$ and sparsifies it, resulting in the predictive subgraph $\Gcal_s$, which retains only the relevant information for the task, i.e., the motifs. Finally, the graph classifier $\theta$ takes $\Gcal_s$ as input and produces the prediction $y \in \{0, 1\}$. }
  \label{fig:pipeline}
\end{figure}

This viewpoint of interpretability is exemplified in \cref{fig:pipeline} with the synthetic \bashapes dataset \cite{GNNExplainer2019}. 
In this dataset, a binary classification task can be solved using only specific motifs of the graphs (the house or the cycle motifs, i.e., yellow nodes), while the original graphs contain additional irrelevant information for the task at hand, i.e., purple nodes.
A sparser graph reduces the volume of information used for making predictions, rendering it easier for humans to understand the GNN predictions. It is crucial to note that interpretability is achieved through sparsity only when the subgraph completely excludes information from the omitted nodes and edges since only then the subgraph, and thus, the explanation is \emph{faithful} to the model prediction.
This situation does not arise, for example, if message passing  \cite{Kipf2016SemiSupervisedCW} is applied to obtain the node embeddings before finding the subgraph.

While most of explainability methods for GNNs are post-hoc, some existing approaches attempt to identify the predictive subgraph during training \cite{Cangea2018TowardsSH, SUGAR}. 
These methods, however, exhibit two primary limitations: 
i) they often still rely on the entire graph for predictions \cite{Cangea2018TowardsSH}, resulting in a lack of \emph{faithfulness} to the model prediction (as described above), and/or 
ii) they impose strong assumptions concerning the structure of the predictive subgraph, such as a pre-defined size.
In practical scenarios, relaxing assumptions on the subgraph structure is desirable.
For instance, in community detection within social networks \cite{Contisciani2020CommunityDW, Shchur2018OverlappingCD}, it is crucial to only remove edges connecting users across distinct communities but the number of inter-community connections varies significantly and is unknown a priori.
Another relevant example is the prediction of mutagenic effects in compounds, where \citet{Debnath1991StructureactivityRO} observed that diverse ``motifs'' of chemical elements are predictive. We refer the reader to \cref{sec:experiments} for a concrete example on the \mutag\ dataset.

In this work, we introduce \ours, a novel algorithm for training GNNs that addresses the two fundamental challenges simultaneously:
i) finding the informative subgraph $\Gcal_s$ by removing edges and/or nodes, \emph{without imposing assumptions on the subgraph's structure}, and 
ii) optimizing the performance of the graph classification task solely using the selected subgraph $\Gcal_s$, thereby enhancing the interpretability of the classifier.
To this end, we use a bi-level optimization approach. 
The optimization for performance adheres to the conventional supervised problem paradigm. 
We then apply reinforcement learning to identify the predictive subgraph, using a policy that allows edge and/or node removal modes.
Our carefully designed reward function allows us to introduce inductive biases toward either sparse or high-performing solutions and uses conformal predictions \cite{Angelopoulos2021AGI} to account for classifier uncertainty at each training epoch.

In summary, our work presents the following key contributions:
(i) A novel framework, \ours, which enables graph-level classification using only predictive subgraphs as input,  leading to more interpretable solutions.
(ii) The flexibility to choose between achieving sparsity through node or edge removal, which removes assumptions on the structure of the predictive subgraph and broadens the range of application scenarios.
(iii) The design of a versatile reward function that takes into account the uncertainty of the classifier and allows to introduce inductive biases towards sparser or more high-performing solutions.
(iv) A comprehensive comparison across nine graph classification datasets, showing that \ours\ makes predictions using significantly sparser graphs while keeping competitive performance.

	\section{Preliminaries} 
\label{sec:background}
This section outlines necessary background information on the key building blocks of \ours: reinforcement learning, message-passing graph neural networks, and conformal predictions. We begin by introducing the notation used throughout this work.

\myparagraph{Notation}

A graph is denoted as $\Gcal = (\Vcal, \Ecal)$ where $\Vcal \in \{1, \dots, n\} = [n]$ is the set of $n$ nodes and $\Ecal \subseteq \Vcal \times \Vcal$ is the set of edges. 
The size of a set is represented as $\size{\Vcal}$.
A subgraph of $\Gcal$ is denoted as $\Gcal[s] = (\Vcal[s], \Ecal[s]) \subseteq \Gcal$ comprising subsets of nodes $\Vcal[s] \subseteq \Vcal$ and edges $\Ecal[s] \subset \Ecal$.
A labeled dataset is represented as $\Dcal = \{\Gcal_i, \yb_i \}_i$ where $\yb_i$ denotes the target of $\Gcal_i$. Here, without loss of generality, we focus on $K$ classes graph classification, hence $\yb_i \in \{1, \dots, K\} = [K]$.

\subsection{Reinforcement learning}

Reinforcement Learning (RL) is a learning paradigm in which an agent learns to optimize decisions by interacting with an environment \cite{Sutton2005ReinforcementLA}. 
The objective is to find a policy $\pi$ that maximizes the expected cumulative reward. 
An RL problem is typically modeled as a Markov Decision Process (MDP), defined by a tuple $(\mathcal{S}, \mathcal{A}, \mathcal{P}, \reward, \gamma)$, where $\mathcal{S}$ and $\mathcal{A}$ denote the state and action spaces, $\mathcal{P}$ the transition probability, $\reward$ the reward function, and $\gamma$ the discount factor.

\myparagraph{Policy Gradient Methods}
Policy gradient methods directly optimize the policy using gradient ascent on the expected cumulative reward \cite{Sutton1999PolicyGM}. 
Proximal Policy Optimization (PPO) \cite{schulman2017proximal} is a policy gradient method that introduces an objective function fostering exploration while mitigating drastic policy updates. PPO aims to solve the optimization problem:

\begin{equation}
L^{CLIP}(\phi) = \mathbb{E}_{t}\left[\min\left(r_t(\phi)\hat{A}_t, clip(r_t(\phi), 1-\epsilon, 1+\epsilon)\hat{A}_t\right)\right]
\label{eq:ppo}
\end{equation}

where $r_t(\phi) = \frac{\pi_{\phi}(a_t|s_t)}{\pi_{\phi_{old}}(a_t|s_t)}$ is the probability ratio, $\hat{A}_t$ an estimator of the advantage function at time $t$, and $\epsilon$ a hyperparameter controlling the deviation from the old policy.

\subsection{Graph neural networks}
Message passing Graph Neural Networks (GNNs) are designed for graph-structured data processing. Each node $v \in \Vcal$ has a feature vector $\hb$, and the GNN transforms these features using neighborhood information.
Formally, a GNN performs $L$ update rounds, updating each node's feature vector $\hb_i^{(l)}$ at step $l \geq 1$:

\begin{equation}
    \hb_i^{(l)} = \frm_{\theta_u} \left( \bigoplus_{j \in \Ncal_i } \mrm_{\theta_m} ( \hb_i^{(l-1)}, \hb_j^{(l-1)}) \right).
\end{equation}

Here, $ \frm_{\theta_u}$ and $\mrm_{\theta_m}$ are differentiable, parameterized functions, $\Ncal_i$ represents node $i$'s neighbors, and $\bigoplus$ denotes permutation invariant operations, i.e., sum, mean, or max. After $L$ steps, we obtain the final node representations ${\hb_i^{(L)}} \forall i \in \Vcal$. These representations are used for tasks like graph-level prediction, employing a permutation-invariant readout function $\hat{\yb} = \grm\left(\left\{\hb_i^{(L)}: i \in \Vcal\right\}\right)$ that ensures the output is node order independent.

\subsection{Conformal Prediction}

Conformal prediction \cite{Angelopoulos2021AGI} is a framework that rigorously quantifies uncertainty in machine learning predictions. 
Given a labeled dataset $\{x_i, y_i\}_{i=1}^N$, a heuristic measure of uncertainty of our predictor (e.g., softmax values from a classifier $\frm_\theta$), and a scoring function $s(x, y) \in \mathbb{R}$ that reflects prediction uncertainty, conformal prediction generates a prediction set for a new input $x_{\text{test}}$ as follows:

\begin{equation}
    \mathcal{C}\left(x_{\text{test}} \right)=\left\{y: s\left(x_{\text{test}}, y\right) \leq \hat{q}\right\} \subseteq [K].
    \label{eq:prediction_set}
\end{equation}

Here, $\hat{q}$ represents the $\frac{\lceil(n+1)(1-\alpha)\rceil}{n}$ quantile of the calibration scores $\{s_i\}_n$, where $\alpha$ is a predefined error rate. 
In the context of classification tasks, one commonly used conformal procedure is Adaptive Prediction Sets (APS) \cite{Angelopoulos2020UncertaintySF,Romano2020ClassificationWV}, which defines the scoring function as:

\begin{equation}
    s(x, y)=\sum_{j=1}^k \frm_\theta(x)_{\pi_j(x)} \text {, where } y=\pi_k(x)
    \label{eq:aps_score}
\end{equation}

Here, $\pi(x)$ represents the permutation of $[K]$ that arranges the softmax values $\frm_\theta(x)$ from most likely to least likely.

	\section{Related work} 
\label{sec:relatd_work}

 GNNs extend deep learning models to incorporate graph-structured data \cite{Bronstein2016GeometricDL}. Numerous architecture proposals have emerged, tailored for node, link, and graph prediction tasks \cite{Scarselli2009TheGN, Gilmer2017NeuralMP, Wu2019ACS, Wu2020GraphNN, GIN, Corso2020PrincipalNA,Chen2020SimpleAD}. 
 However, these approaches rely on the complete graph data and often disregard interpretability, which is the focus of our work.
More recently, some works aim for sparsification \cite{zheng2020robust,DropEdge_SmarterRemoval,Rong2019DropEdgeTD,Oversmoothing,Oversmoothing2,SGCN,PTDNET,SparRL,wang2021bi}.
However, these methods typically make strong assumptions on the predictive subgraph. Moreover, they often focus on tasks other than graph prediction.

Here, we explore relevant sparsification techniques that specifically target supervised problems, which are the main focus of our paper.
We categorize these works into two categories: those employing soft information removal, where the predictive subgraph retains information from the complete graph, and thus may not be faithful to the model predictions; and those adopting hard information removal, which closely aligns with our approach.

\myparagraph{Soft removal of information} 
Several architectures \cite{Javaloy2022LearnableGC, Velickovic2017GraphAN, Brody2021HowAA} assign varying importance to network edges, simulating soft edge removal whilst maintaining information flow across nodes. 
\diffpool\ \cite{diffpool} hierarchically removes parts of the original graph, distilling the graph's representation into a single vector used for prediction. 
\topksoft\ \cite{Lee2019SelfAttentionGP, Knyazev2019UnderstandingAA}, uses a GNN to update node features and subsequently selects the top $k$ most relevant nodes based on their features. It is worth noting that even though part of the network is dropped, the remaining subgraph contains information from the entire network. In contrast, in our framework, the graph predictor does not have access to the dropped information.

\myparagraph{Hard removal of information}
\topkhard\ \cite{Gao2019GraphU,Cangea2018TowardsSH, Knyazev2019UnderstandingAA} selects the top $k$ nodes based solely on their information. However, the number of nodes is fixed and cannot dynamically change for different graphs.
\sugar\ \cite{SUGAR} takes a different approach by identifying discriminative subgraphs. It introduces a reinforcement pooling module trained with Q-learning \cite{Mnih2015HumanlevelCT} to adaptively select a pooling ratio when recombining subgraphs for classification. 
SUGAR also imposes a predefined size for the subgraphs. 
In contrast, \ours\ does not make any assumptions about the structure of the subgraph, uses policy gradient methods to capture uncertainty in the sparsity process, and allows for different modes of removal.

Finally, there is extensive literature that aims to understand which subgraph of the input graph is most relevant for making predictions in GNNs \cite{GNNExplainer2019, PGExplainer2020, yuan2020xgnn, numeroso2021meg, Yuan2021SubgraphX, yu2022motifexplainer, GraphMask21}. 
However, it is important to note that these methods primarily focus on explaining a GNN predictor trained using the original graphs. In contrast, our objective is to train a GNN that inherently utilizes a minimal portion of the original graph.
While these methods may complement our approach by providing further insights of the predictions, they operate in a post-hoc manner.

	\section{\ours: Conformal-based reinforcement learning for graph sparsification }
\label{sec:method}

In this section, we present \ours, a novel training procedure for GNNs leveraging reinforcement learning with conformal-based rewards to achieve graph sparsification. 
Given a labeled dataset $\Dcal$, the main goal of \ours\ is to identify a compact predictive subgraph $\Gcal[s] \subseteq \Gcal$ that maintains high performance for a graph classification task.
Next, we provide a detailed description of the different parts comprising \ours.

\subsection{Optimizing for sparsity and performance}

Firstly, we aim to learn a function $\policy_\phi: \Gcal \mapsto \Gcal[s]$ with parameters $\phi$ responsible for identifying the predictive subgraph. This function corresponds to the policy of the reinforcement learning component of \ours.
Secondly, we want to train a graph classifier $\frm_\theta: \Gcal[s] \rightarrow \hat{\yb}$ with parameters $\theta$ that takes the identified predictive subgraph as input and generates predictions.
The pipeline of our approach is illustrated with the synthetic \bashapes\ dataset in \cref{fig:pipeline}. 
To tackle this challenge we introduce a bi-level iterative optimization approach:

\begin{align}
\phi^{*} &= \myargmin{\phi}{\Lcals}{\theta^{\star}(\phi), \phi, \Dcal[val] }  \label{eq:outer} \\
& \text { s.t. } \theta^\star(\phi) = \myargmin{\theta}{\Lcalp}{\theta, \phi, \Dcal[tr] }  \label{eq:inner}
\end{align}

The nested structure of the problem implies that achieving an optimal predictive subgraph requires a high-performing graph classifier. 
We employ gradient descent for both optimizations, with a larger learning rate for the inner optimization  \cite{Zheng2021StackelbergAG}. 
Also, notice that we use the training $\Dcal[tr]$ and validation $\Dcal[val]$ sets to solve the inner and outer procedures respectively, which avoids overfitting.
\cref{alg:training} summarizes the training procedure.

\myparagraph{Performance optimization} 
This objective corresponds to \cref{eq:inner} which represents a standard graph-supervised problem. Here, the objective is to minimize a loss function $\Lcalp$, e.g., the cross-entropy loss. Thus the goal is to learn a function $\frm_\theta: \Gcal[s] \rightarrow \hat{\yb}$ with parameters $\theta$ that minimizes the prediction loss.

\begin{algorithm}[t]
\caption{Training of \ours.}
\label{alg:training}
\begin{algorithmic}[1]
\State \textbf{Input:} Training $\Dcal[tr]$ and validation $\Dcal[val]$ sets; Initial graph classifier parameters $\theta$ and policy parameters $\phi$; Empty buffer $\Bcal$; Learning rates $\alpha$ and $\beta$; Number of PPO updates $K$
\While{not convergence}
    \For{$\{\yb, \Gcal\} \in \Dcal[tr]$}  \Comment{Get a (batch of) sample(s) from the training set}
        \State $\policy_\phi: \Gcal \mapsto \Gcal[s]$ \Comment{Get the subgraph using the policy}
        \State $\hat{\yb}_s = \frm_{\theta}(\Gcal_s)$ \Comment{Get the prediction }
        \State $\theta \leftarrow \theta - \alpha \nabla_\theta \Lcalp(\theta, \phi)$ \Comment{Update the parameters of the graph classifier}
    \EndFor
    \For{$\{\yb, \Gcal \} \in \Dcal[val]$}  \Comment{Get a (batch of) sample(s) from the validation set}
        \State  $\policy_\phi: \Gcal \mapsto \Gcal[s]$  \Comment{Use the $\policy$ to sample a sparse graph}
        \State $\yhat_s = \frm (\Gcal[s])$ \Comment{Get the prediction of the sparse graph}
        \State Add tuple $\{\yb, \Gcal, \yhat, \Gcal[s] \}$ to the buffer $\Bcal$
    \EndFor
    \State Compute the quantile $\hat{q}$ using the calibration scores obtained with $\Dcal[tr]$
    \State Compute rewards and prepare mini-batches from buffer $\Bcal$ for PPO updates
    \For{i = $\{1, 2, \dots, K\}$} \Comment{Update PPO K times}
        \State $\{\yb, \Gcal, \yhat, \Gcal[s], r\} \sim \Bcal$\Comment{Sample a batch of tuples from the buffer}
        \State Compute advantage estimates $\hat{A}$
        \State   $\phi \leftarrow \phi - \beta \nabla_\phi \Lcals(\theta, \phi)$ \Comment{Update policy by maximizing  \cref{eq:ppo}}
    \EndFor
    \State Check for convergence, update convergence flag
\EndWhile
\State \textbf{Output:} Optimal policy parameters $\phi$ and graph classifier parameters $\theta$
\end{algorithmic}
\end{algorithm}

\myparagraph{Sparsity optimization }
This optimization task corresponds \cref{eq:outer}. 
It presents a considerable challenge due to the combinatorial nature of the node and edge removal process.
Specifically, we must determine both which edges ($(u, v) \in \Ecal$) and/or nodes  ($v \in \Vcal$) and how many of them, should be removed.
Drawing inspiration from SparRL \cite{SparRL}, we formulate the sparsification task as a Markov Decision Process (MDP) and address it through the framework of graph reinforcement learning \cite{Nie2022ReinforcementLO}: we parameterize the policy $\policy_\phi$ using a GNN. 
In contrast to prior approaches that rely on value-based methods like Deep Q-Learning \cite{Mnih2015HumanlevelCT}, we opt for the policy gradient method Proximal Policy Optimization (PPO) \cite{PPO} to capture the inherent uncertainty in the sparsification process. 
We denote the objective of sparsity optimization as $\Lcals$, which corresponds to solving the PPO objective in \cref{eq:ppo}. 
The outcome of this task is a policy $\policy_\phi$ that finds the predictive subgraph $\Gcal[s]$, subsequently employed as input for the graph classification task.
Next, we provide more details on the sparsity optimization.

\subsection{Unpacking the sparsity optimization}

In this subsection, we provide a detailed description of the components of the PPO method we designed to identify the predictive subgraph. 
In RL problems, agents usually make sequential decisions over multiple time steps, and the rewards they receive at each step may depend on the whole trajectory taken so far. 
This challenge, known as the ``delayed reward''  problem, is a central challenge in RL \cite{sutton1992introduction}.
Our objective shifts towards identifying the predictive subgraph in a single step, rather than considering a trajectory as in traditional RL. 
This approach aims to find a strategy that maximizes the reward across the dataset graphs.
This simplified scenario resembles the Multi-Armed Bandit Problem \cite{kuleshov2014algorithms}.
As our experiments in \cref{sec:experiments} empirically show, this simplified scenario effectively achieves high-performance relying on significantly sparser graphs.
Below, we focus on describing two key components: the policy responsible for the removal process and the design of the reward function that guides the policy in finding the optimal predictive subgraph.

\subsubsection{Policy formulation for graph sparsification via node or edge removal}
\label{sec:policy}

We propose a policy $\policy[\ab | \Gcal; \phi]$ we model using a GNN with parameters $\phi$. 
Given an input graph $\Gcal$, the policy produces an action $\ab$, where each element determines whether to keep ($a_i = 0$) or remove ($a_i = 1$) a specific node or edge in $\Vcal$ or $\Ecal$, respectively. 
This action results in a subgraph $\Gcal[s] \subseteq \Gcal$.
Recognizing the inherent uncertainty in real-world scenarios where the importance of a node/edge may not be clear, we design the policy as a probability distribution. 
We establish two operational modes depending on the removal objective:

\begin{itemize}
\item \textbf{Node Removal Policy}: $\policyn[\ab | \Gcal; \phi]$, where $\ab \in \{0, 1\}^{\size{\Vcal}}$. In this scenario, the resulting subgraph $\Gcal[s]$, used as input for the downstream task , satisfies $\Vcal[s] = \{v \in \Vcal | a_v = 0 \}$ and $\Ecal[s] = \{(u, v) \in \Ecal | a_u = 0 \land a_v=0 \}$.
\item \textbf{Edge Removal Policy}: $\policye[\ab | \Gcal; \phi]$, where $\ab \in \{0, 1\}^{\size{\Ecal}}$. Here, the subgraph $\Gcal[s]$, used as input for the downstream task, has $\Vcal[s] = \Vcal$ and $\Ecal[s] = \{(u, v) \in \Ecal | a_{uv}=0 \}$.
\end{itemize}

Note that node removal implies the removal of all edges connected to the node.
In contrast, the edge removal policy retains all nodes, making it suitable for node classification, which is a direction for future work.
In either case, determining $\Gcal[s]$ poses an NP-hard problem for both removal modes. 
The number of potential subgraphs increases exponentially with the number of nodes/edges in the original graph. 
Specifically, the number of possible actions for $\policyn$ is $\sum_{i=1}^{\size{\Vcal}-1} \binom{\size{\Vcal}}{i}$, while for $\policye$, it is $\sum_{i=1}^{\size{\Ecal}} \binom{\size{\Ecal}}{i}$. 
RL can be helpful to solve combinatorial problems, particularly when the reward function effectively guides the optimization process \cite{Mazyavkina2020ReinforcementLF}.

\subsubsection{Reward formulation}
\label{sub:reward}

The reward function $\reward$ plays a pivotal role in shaping the policy's behavior. 
It should provide positive rewards for predictive and sparse subgraphs while penalizing those that negatively impact the performance of the graph classifier. 
It is also important to consider that the classifier makes errors, i.e., it does not achieve perfect performance. 
Consequently, there are instances where a subgraph may be genuinely predictive, but the classifier produces an incorrect prediction. 
Our reward design accounts for this inherent uncertainty through the use of conformal predictions, as described in a subsequent section. It consists of two primary components: one aimed at enhancing performance and the other dedicated to promoting sparsity.

\myparagraph{Performance Reward ($\rewardp$)} Our approach to rewarding performance is straightforward. We simply use the softmax score assigned by the graph classifier to the correct class $y$, i.e., $\rewardp=\frm_\theta(\Gcal_s)_{y} \in [0, 1]$.

\begin{wrapfigure}[18]{r}{0.43\textwidth}
  \centering
  \includegraphics[width=0.4\textwidth]{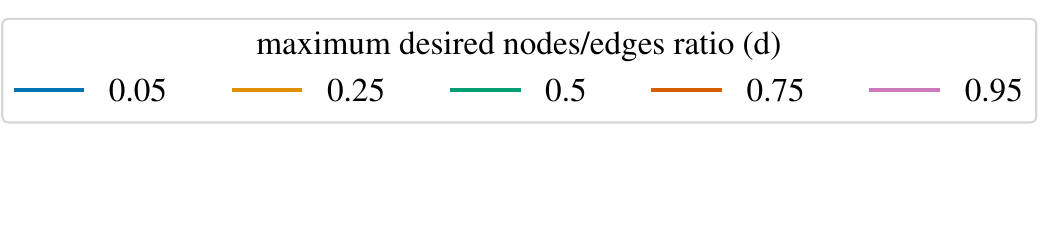}  
  \begin{subfigure}[c]{0.4\textwidth}
        \centering
        \includegraphics[width=\textwidth]{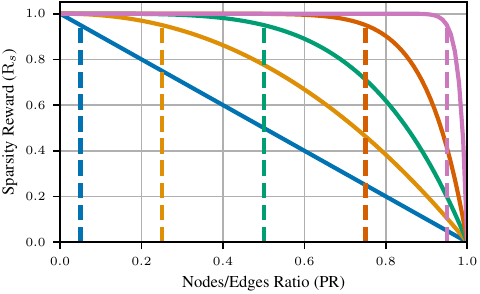}
    \end{subfigure}
    \caption{Evolution of the sparsity reward $\rewards$ over the node/edge ratio for different values of the \textit{maximum desired nodes/edges ratio} $d$.} 
\end{wrapfigure}

\myparagraph{Sparsity Reward ($\rewards$)} 
To incentivize a minimal predictive subgraph, we introduce a component that penalizes the \textit{nodes ratio} $\PR[n]=\frac{\size{\Vcal[s]}}{\size{\Vcal}}$ or the \textit{edges ratio} $\PR[e]=\frac{\size{\Ecal[s]}}{\size{\Ecal}}$ kept from the original graph, depending on the policy mode. %
To control the desired sparsity level, we introduce a parameter $d \in [0,1]$ representing the \emph{maximum desired nodes/edges ratio}.
Then, we define the sparsity reward as $\rewards(\Gcal[s]) = 1 - \PR^{\tilde{d}}$, where $\tilde{d}$ is a transformation of $d$ such that $\rewards=0.95$ when $\PR=d$. 
The variation of $\rewards$ with respect to $\PR$ for different $d$ values is depicted in the inline figure. 
Each vertical line represents a unique $d$ value, extending upwards until $\rewards = 0.95$ for all instances.

To seek sparsity, a smaller $d$ value (e.g., blue or yellow curves) is preferred. 
This leads to a slower increase in $\rewards$ with node/edge removal, rewarding substantial removal. 
Conversely, when $d$ approaches 1 (e.g., pink curve), $\rewards$ close to 1 are awarded even for keeping most nodes/edges.

Note that both reward components fall within the range $[0,1]$, facilitating comparison.
Now, we proceed to define the reward function used in \ours:

\begin{equation}
 \reward = 
\begin{cases}
    \lambda \rewardp +  (1 - \lambda)\rewards(\Gcal_s)  & \text{if } y  \in \cset[\Gcal_s]  \land \size{\cset[\Gcal_s]} = 1, \\
    \frac{\rewardp}{\size{\cset[\Gcal_s]}}  & \text{if } y  \in \cset[\Gcal_s] \land \size{\cset[\Gcal_s]} > 1, \\
     - \rewards(\Gcal_s) & \text{if } y \not\in \cset[\Gcal_s].
\label{eq:reward}
\end{cases}
\end{equation}

There are two key aspects to highlight. 
Firstly, we compute the quantile $\hat{q}$ used to obtain the prediction set $\cset$ using the scoring function introduced in \cref{eq:aps_score}. 
Secondly, it involves three different cases based on whether the true label $y$ is within the prediction set and the size of the prediction set.

The top scenario corresponds to cases in which the predictive subgraph leads to a highly certain and correct prediction. 
Here, a parameter $\lambda \in [0, 1]$ allows practitioners to emphasize either performance ($\lambda = 1$) or sparsity ($\lambda = 0$). 
In this scenario, the reward is in the range $\Rrm \in [0, 1]$.
The middle scenario addresses instances in which the subgraph leads to the classifier being uncertain about the prediction, i.e., $\size{\cset[\Gcal_s]} > 1$. 
In such cases, we aim to provide a positive but smaller reward within the range $(0, 0.5)$. 
This reward encourages to increase the probability of the true label and to reduce the size of the prediction set, that is moving to the top scenario. Intuitively, this scenario arises with challenging examples.
The bottom scenario refers to cases in which the predictive subgraph results in an incorrect prediction, falling outside the prediction set $\cset$. 
As this behavior is undesirable, the rewards $\Rrm$ fall within the range $[-1, 0]$, encouraging to reduce sparsity. In this scenario, the reward reaches 0 only when we recover the original graph.

The use of conformal prediction in the reward function is essential to guide the policy toward discovering a good predictive subgraph. If the task is challenging and the classifier exhibits uncertainty, the focus should initially be on performance. Only when the classifier is confident about the prediction should we encourage sparsity. 
In the following section, we present empirical evidence supporting the reward function's design, and more generally our approach \ours, for achieving both performance and compact predictive subgraphs.

	\section{Experiments} 
\label{sec:experiments}
In this section, we present a comprehensive set of experiments to assess the performance of \ours.
First, we conduct an ablation study to show the impact of different values of $\lambda$ and maximum desired ratio ($d$) on performance, as well as the effect of the choice of base GNN architecture.
Then, we compare \ours\ with relevant baselines.

\myparagraph{Proposed approaches} We evaluate the two information removal modes of the proposed framework, described in \cref{sec:policy}. We denote the model with the policy that removes nodes as \ours[N] and the model with the policy that exclusively removes edges as \ours[E].

\myparagraph{Datasets}
 In this study, we used seven Bioinformatics datasets, which include \mutag\ \cite{mutag}, \dd\ \cite{dd}, \enzymes\ \cite{enzymes}, \ncione\ \cite{ncidataset}, \nciten\ \cite{ncidataset}, \ptc\ \cite{ptc}, and \proteins\ \cite{enzymes}. Additionally, we incorporated two chemical compound datasets, \bzr\ \cite{Fey/Lenssen/2019} and \cox\ \cite{Fey/Lenssen/2019}. \cref{tab:dataset-stats} shows the statistics of the datasets.

\begin{table}[t]
\caption{\textbf{Statistics of the datasets.}}
\centering
{\small
\begin{tabular}{lrrrrrrr}
\toprule
Dataset & \# Graphs & \# Features & \# Edge Features & \# Classes & Undirected & \# Nodes & \# Edges \\
\midrule
BZR & 405 & 10 & 0 & 2 & Yes & 35.75 $\pm$ 7.26 & 76.72 $\pm$ 15.39 \\
COX2 & 467 & 10 & 0 & 2 & Yes & 41.22 $\pm$ 4.03 & 86.89 $\pm$ 8.53 \\
DD & 1178 & 20 & 0 & 2 & Yes & 284.32 $\pm$ 272.00 & 1431.32 $\pm$ 1387.81 \\
ENZYMES & 600 & 21 & 0 & 6 & Yes & 32.63 $\pm$ 15.28 & 124.27 $\pm$ 51.00 \\
MUTAG & 135 & 7 & 4 & 2 & Yes & 18.85 $\pm$ 4.49 & 41.67 $\pm$ 11.13 \\
NCI1 & 4110 & 7 & 0 & 2 & Yes & 29.87 $\pm$ 13.56 & 64.60 $\pm$ 29.87 \\
NCI109 & 4127 & 8 & 0 & 2 & Yes & 29.68 $\pm$ 13.57 & 64.26 $\pm$ 29.92 \\
PROTEINS & 1113 & 4 & 0 & 2 & Yes & 39.06 $\pm$ 45.76 & 145.63 $\pm$ 169.20 \\
PTC & 349 & 8 & 4 & 2 & Yes & 14.11 $\pm$ 8.44 & 28.97 $\pm$ 18.88 \\
\bottomrule
\end{tabular}
}
\label{tab:dataset-stats}
\end{table}

\myparagraph{Experimental setup} 
Aiming for computational efficiency, we conducted cross-validation on the hyperparameters of the vanilla GNN classifier, i.e., the GNN architecture (e.g., \gin) trained with stochastic gradient descent. We selected the optimal configuration of hyperparameters based on the validation set and used it to train the remaining models.  We performed 5-fold cross-validation and reported standard deviation values on the test set at the last epoch. All experiments were conducted on a single CPU with 8GB of RAM. 
Here, we present results obtained using the \gin\ architecture. 
Additional results using other well-known architectures, namely \gat\ and \gcn, are provided in \cref{app:comparison}. For a comprehensive overview of the best configuration hyperparameters obtained for each model, architecture and dataset, please refer to \cref{app:training}.
We implemented the baselines using the Torch Geometric package \cite{torchgeometric} whenever it was available. Furthermore, we provide the implementation of \ours, along with the necessary scripts to replicate the experiments, at \url{https://github.com/psanch21/CORES}.

\subsection{Ablation study}

In this section, we provide an ablation study on the proposed framework analyzing the effects of key hyperparameters and different choices for the base GNN architecture.

\begin{figure}[tbp]
	\centering
	\includegraphics[width=0.215\textwidth]{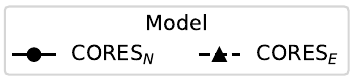}
	\includegraphics[width=0.4\textwidth]{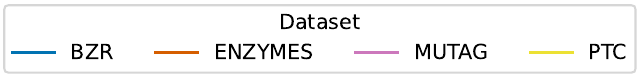}
	
	\begin{minipage}{0.49\textwidth}
		\centering
		\begin{subfigure}[b]{1.0\textwidth}
			\begin{subfigure}[b]{0.49\textwidth}
				\includegraphics[width=\textwidth]{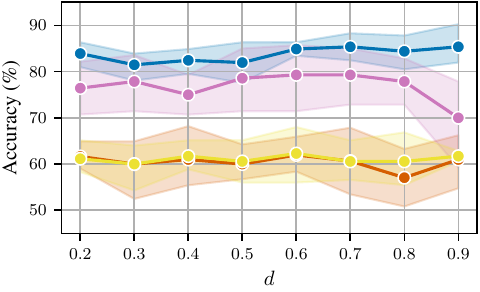}
			\end{subfigure}
			\begin{subfigure}[b]{0.49\textwidth}
				\includegraphics[width=\textwidth]{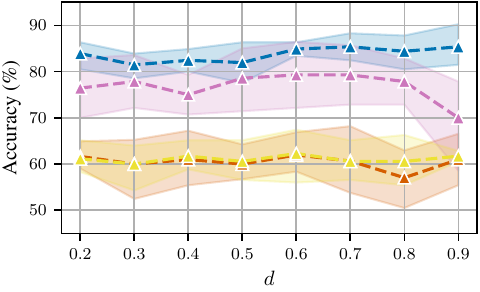}
			\end{subfigure}
			\caption{$d$ versus accuracy}
			\label{fig:d_perf}
		\end{subfigure}
		\begin{subfigure}[b]{1.0\textwidth}
			\begin{subfigure}[b]{0.49\textwidth}
				\includegraphics[width=\textwidth]{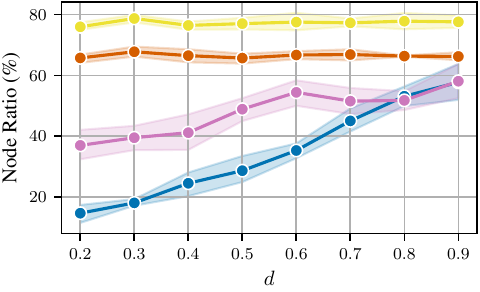}
			\end{subfigure}
			\begin{subfigure}[b]{0.49\textwidth}
				\includegraphics[width=\textwidth]{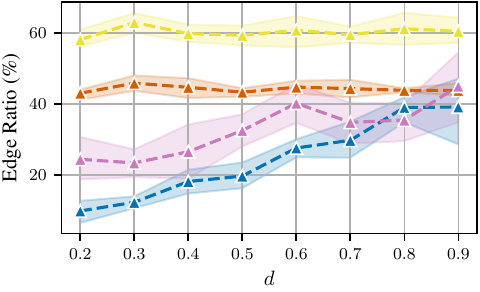}
			\end{subfigure}
			\caption{$d$ versus sparsity}
			\label{fig:d_spar}
		\end{subfigure}
		\begin{subfigure}[b]{1.0\textwidth}
			\begin{subfigure}[b]{0.49\textwidth}
				\includegraphics[width=\textwidth]{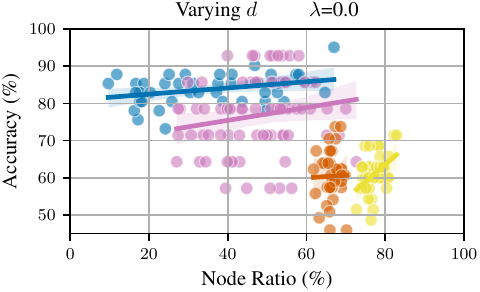}
			\end{subfigure}
			\begin{subfigure}[b]{0.49\textwidth}
				\includegraphics[width=\textwidth]{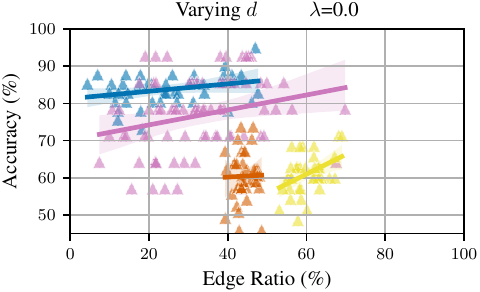}
			\end{subfigure}
			\caption{Sparsity versus accuracy}
		\end{subfigure}
		\caption{\textbf{Ablation study on the maximum desired ratio $d$.} We use the \gin\ architecture and run each experiment for 5 different folds.}
		\label{fig:ablation_d}
	\end{minipage}
	\hfill
	\begin{minipage}{0.49\textwidth}
		\centering
		\vspace{-0.3cm}
		\begin{subfigure}[b]{1.0\textwidth}
			\begin{subfigure}[b]{0.49\textwidth}
				\includegraphics[width=\textwidth]{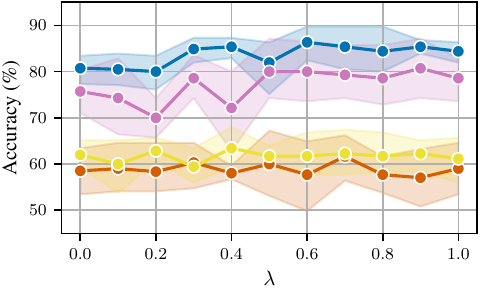}
			\end{subfigure}
			\begin{subfigure}[b]{0.49\textwidth}
				\includegraphics[width=\textwidth]{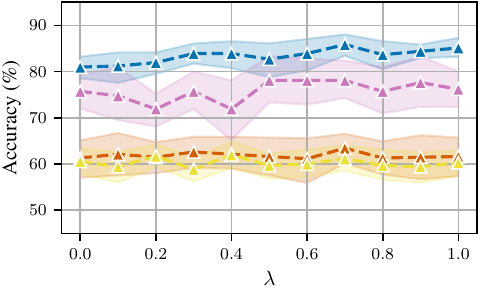}
			\end{subfigure}
			\caption{$\lambda$ versus accuracy}
		\end{subfigure}
		\begin{subfigure}[b]{1.0\textwidth}
			\begin{subfigure}[b]{0.49\textwidth}
				\includegraphics[width=\textwidth]{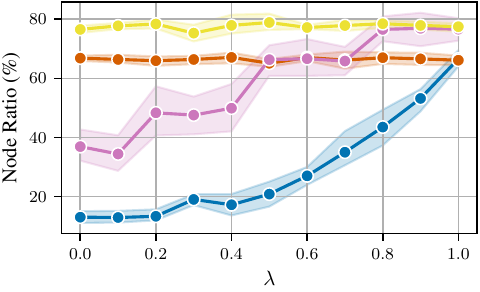}
			\end{subfigure}
			\begin{subfigure}[b]{0.49\textwidth}
				\includegraphics[width=\textwidth]{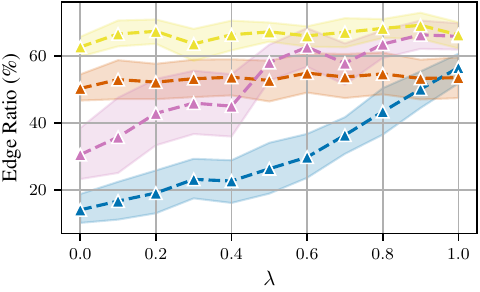}
			\end{subfigure}
			\caption{$\lambda$ versus sparsity}
		\end{subfigure}
		\begin{subfigure}[b]{1.0\textwidth}
			\begin{subfigure}[b]{0.49\textwidth}
				\includegraphics[width=\textwidth]{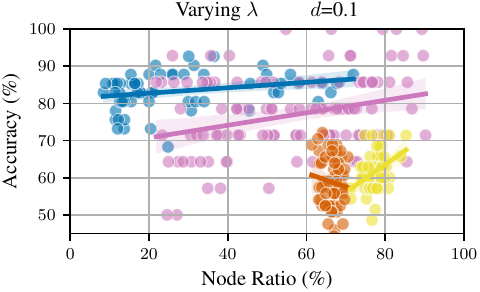}
			\end{subfigure}
			\begin{subfigure}[b]{0.49\textwidth}
				\includegraphics[width=\textwidth]{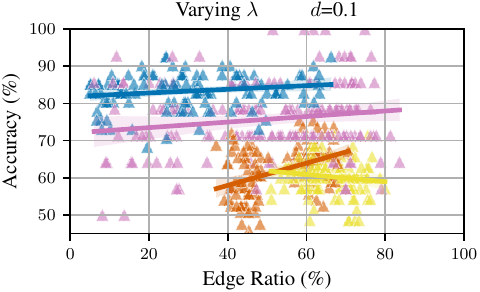}
			\end{subfigure}
			\caption{Sparsity versus accuracy}
		\end{subfigure}
		\caption{\textbf{Ablation study on $\lambda$.} We use the \gin\ architecture and run each experiment for 5 different folds.}
		\label{fig:ablation_lambda}
	\end{minipage}
\end{figure}

\myparagraph{Analysis of the impact of $\lambda$ and $d$}
The reward function, presented in \cref{eq:reward}, leverages two hyperparameters: the maximum desired nodes/edges ratio ($\desired$)  and the balancing parameter ($\lambda$) to prioritize performance or sparsity.  
For the nine datasets under study, we present the effect of varying $\desired$ and $\lambda$ in \cref{fig:ablation_d} and \cref{fig:ablation_lambda}, respectively, for both \ours[N] (solid lines) and \ours[E] (dashed lines). For this ablation study, we use \gin\ as the base GNN architecture. The results are color-coded for each dataset.
The figures in the top row contain the performance analysis, while the middle row shows the sparsity, specifically the node ratio ($\frac{\size{\Vcal_s}}{\size{\Vcal}}$) for \ours[N] and the edge ratio ($\frac{\size{\Ecal_s}}{\size{\Ecal}}$) for \ours[E].
The bottom row shows performance versus sparsity in the vertical and horizontal axes respectively.

In \cref{fig:ablation_d}, we fix $\lambda=0.0$ (only rewarding sparsity) while varying $\desired$ within the range $[0.2, 0.9]$.
Regarding performance (top row), we observe that generally, the value of  $\desired$ does not have an impact. 
The middle row, which focuses on sparsity, reveals interesting variations among datasets. 
For instance, for the \mutag\ dataset, \ours\ finds a predictive subgraph that utilizes as little as 20\% of the original nodes/edges with $\desired=0.2$. 
In contrast, for the \ptc\ dataset, we keep approximately 80\% of the original nodes in \ours[N] and 60\% of the original edges in \ours[E], even with low $\desired$ values.
The bottom figures highlight that for certain datasets, e.g., \bzr, \ours\ can maintain comparable accuracy regardless of sparsity levels. Conversely, for other datasets, decreasing sparsity tends to enhance performance, such as the case with the \ptc\ dataset and \ours[N].
These findings suggest that some datasets exhibit rapid performance degradation when information is removed. 
Examining the different reward scenarios in our reward definition (refer to \cref{eq:reward}), it is intuitive to deduce that \ours\ will prioritize reducing sparsity if the prediction is not within the prediction set (bottom scenario) or enhancing performance when the classifier exhibits uncertainty (middle scenario).

In \cref{fig:ablation_lambda}, we fix $\desired=0.1$ (favoring high levels of sparsity) and we vary $\lambda$ within the range $[0.0, 1.0]$.
We observe a similar behavior as in \cref{fig:ablation_d}. 
The top row consistently shows that varying $\lambda$ has a minimal impact on accuracy, and only sometimes leads to increased accuracy as $\lambda$ increases (e.g., \bzr\ dataset). 
Importantly, for none of the datasets does performance degrade with higher $\lambda$ values. 
This behavior aligns with our objective, as higher values of $\lambda$ correspond to a greater emphasis on rewarding performance.
The middle row reveals two distinct patterns: i) a high correlation between $\lambda$ and the number of nodes/edges kept, particularly evident in datasets like \mutag, and ii) minimal sensitivity, as observed with the \ptc\ dataset.
The bottom row reveals similar patterns: positive correlation between the node/edge ratio and accuracy. It is also worth noting that varying $\lambda$ results in a wide range of sparsity levels for some datasets, such as 20-90\% for \mutag, and a narrower range for others, like 40-70\% for \ptc.

In summary, we observe that varying $\desired$ or $\lambda$ leads to one of two scenarios, depending on the dataset: 
i) A minimal change in performance and low sparsity, which suggests that information removal significantly degrades performance for these datasets.
ii) A small positive correlation between the hyperparameters and performance and a high negative correlation between the hyperparameters and sparsity.
Based on these insights, we recommend that practitioners use $\desired$ to fine-tune sparsity levels and consider $\lambda=1.0$ when prioritizing performance over sparsity. \cref{app:ablation-hyper} contains the results for the rest of the datasets under study.

\begin{table}[t]
\caption{\textbf{Ablation study on GNN architecture.} We compare the performance (accuracy) and sparsity (node/edge ratio) of \ours\ with the \vanilla\ approach for the \gcn\ (left) and \gat (right) architectures.  We show the mean over 5 independent runs and the standard deviation as the subindex. All metrics are shown in percentages.}
\centering
\begin{tabular}{lcccc|ccc}
\toprule
\multirow{2}{*}{\textbf{Dataset}} & \multirow{2}{*}{\textbf{Metric}} & \multicolumn{3}{c}{\textbf{\gcn}} & \multicolumn{3}{c}{\textbf{\gat}} \\ 
\cmidrule(lr){3-5}\cmidrule(lr){6-8}

& & \textbf{\vanilla} & \textbf{CORES$_N$} & \textbf{CORES$_E$} & \textbf{\vanilla} & \textbf{CORES$_N$} & \textbf{CORES$_E$}\\ 
\midrule
\multirow{3}{*}{BZR} & \textbf{Accuracy} & 86.83$_{4.08}$ & 82.44$_{4.69}$ & 80.98$_{5.29}$ & 80.98$_{4.36}$ & 83.90$_{4.43}$ & 81.46$_{4.08}$\\ 
 & \textbf{Node Ratio} & - & 76.98$_{6.06}$ & 100.00$_{0.00}$ & - & 37.99$_{10.27}$ & 100.00$_{0.00}$\\ 
 & \textbf{Edge Ratio} & - & 61.17$_{11.31}$ & 46.16$_{5.68}$ & - & 29.92$_{8.83}$ & 57.26$_{21.50}$\\ 
\cmidrule(lr){2-8}
\multirow{3}{*}{COX2} & \textbf{Accuracy} & 79.15$_{4.85}$ & 80.85$_{3.01}$ & 81.70$_{4.90}$ & 82.13$_{4.15}$ & 81.70$_{4.41}$ & 80.85$_{2.61}$\\ 
 & \textbf{Node Ratio} & - & 72.49$_{4.24}$ & 100.00$_{0.00}$ & - & 77.91$_{2.29}$ & 100.00$_{0.00}$\\ 
 & \textbf{Edge Ratio} & - & 53.19$_{8.07}$ & 31.27$_{23.60}$ & - & 62.62$_{3.24}$ & 66.18$_{0.38}$\\ 
\cmidrule(lr){2-8}
\multirow{3}{*}{DD} & \textbf{Accuracy} & 78.47$_{3.47}$ & 79.32$_{2.58}$ & 75.76$_{4.05}$ & 76.10$_{3.02}$ & 75.42$_{4.19}$ & 76.44$_{2.19}$\\ 
 & \textbf{Node Ratio} & - & 56.44$_{2.50}$ & 100.00$_{0.00}$ & - & 75.16$_{1.39}$ & 100.00$_{0.00}$\\ 
 & \textbf{Edge Ratio} & - & 32.01$_{2.89}$ & 51.39$_{1.90}$ & - & 56.42$_{2.16}$ & 58.62$_{2.51}$\\ 
\cmidrule(lr){2-8}
\multirow{3}{*}{ENZYMES} & \textbf{Accuracy} & 66.00$_{5.35}$ & 58.31$_{8.47}$ & 57.02$_{11.62}$ & 71.33$_{10.63}$ & 69.84$_{4.43}$ & 67.54$_{12.23}$\\ 
 & \textbf{Node Ratio} & - & 77.17$_{7.28}$ & 100.00$_{0.00}$ & - & 80.26$_{1.03}$ & 100.00$_{0.00}$\\ 
 & \textbf{Edge Ratio} & - & 62.33$_{9.26}$ & 74.20$_{35.99}$ & - & 65.92$_{1.80}$ & 61.60$_{9.06}$\\ 
\cmidrule(lr){2-8}
\multirow{3}{*}{MUTAG} & \textbf{Accuracy} & 81.43$_{3.91}$ & 81.43$_{10.83}$ & 87.14$_{6.56}$ & 81.43$_{6.39}$ & 80.00$_{5.98}$ & 81.43$_{6.02}$\\ 
 & \textbf{Node Ratio} & - & 79.94$_{2.79}$ & 100.00$_{0.00}$ & - & 75.72$_{4.04}$ & 100.00$_{0.00}$\\ 
 & \textbf{Edge Ratio} & - & 63.52$_{4.23}$ & 32.66$_{8.79}$ & - & 58.74$_{6.05}$ & 55.25$_{14.77}$\\ 
\cmidrule(lr){2-8}
\multirow{3}{*}{NCI1} & \textbf{Accuracy} & 77.33$_{2.09}$ & 66.45$_{7.71}$ & 65.94$_{6.31}$ & 79.42$_{1.34}$ & 73.43$_{2.67}$ & 75.38$_{1.22}$\\ 
 & \textbf{Node Ratio} & - & 76.47$_{17.88}$ & 100.00$_{0.00}$ & - & 63.49$_{3.58}$ & 100.00$_{0.00}$\\ 
 & \textbf{Edge Ratio} & - & 64.02$_{18.84}$ & 37.98$_{14.34}$ & - & 53.33$_{2.81}$ & 88.68$_{6.67}$\\ 
\cmidrule(lr){2-8}
\multirow{3}{*}{NCI109} & \textbf{Accuracy} & 79.32$_{1.57}$ & 76.08$_{2.71}$ & 77.14$_{2.02}$ & 76.90$_{1.34}$ & 74.72$_{2.25}$ & 72.78$_{2.29}$\\ 
 & \textbf{Node Ratio} & - & 78.78$_{4.15}$ & 100.00$_{0.00}$ & - & 71.55$_{11.30}$ & 100.00$_{0.00}$\\ 
 & \textbf{Edge Ratio} & - & 67.08$_{4.42}$ & 92.02$_{1.93}$ & - & 62.37$_{12.96}$ & 62.37$_{13.13}$\\ 
\cmidrule(lr){2-8}
\multirow{3}{*}{PROTEINS} & \textbf{Accuracy} & 72.86$_{1.02}$ & 73.21$_{2.45}$ & 75.89$_{1.94}$ & 75.36$_{3.00}$ & 73.39$_{3.86}$ & 73.57$_{1.62}$\\ 
 & \textbf{Node Ratio} & - & 74.66$_{6.10}$ & 100.00$_{0.00}$ & - & 67.94$_{2.97}$ & 100.00$_{0.00}$\\ 
 & \textbf{Edge Ratio} & - & 56.12$_{9.77}$ & 64.73$_{0.98}$ & - & 46.56$_{4.27}$ & 52.82$_{1.86}$\\ 
\cmidrule(lr){2-8}
\multirow{3}{*}{PTC} & \textbf{Accuracy} & 57.78$_{7.71}$ & 62.86$_{2.86}$ & 63.43$_{3.13}$ & 64.00$_{4.33}$ & 62.96$_{4.17}$ & 59.43$_{4.69}$\\ 
 & \textbf{Node Ratio} & - & 64.10$_{4.19}$ & 100.00$_{0.00}$ & - & 45.61$_{6.00}$ & 100.00$_{0.00}$\\ 
 & \textbf{Edge Ratio} & - & 40.21$_{6.93}$ & 44.47$_{5.29}$ & - & 21.24$_{6.53}$ & 76.36$_{14.75}$\\ 
\bottomrule
\end{tabular}
\label{tab:comparison-archi}
\end{table}
\myparagraph{Ablation on the base GNN architecture} 
Now, we analyze how switching the base architecture affects the performance of \ours. 
\cref{tab:comparison-archi} summarizes the results for two base architectures: \gcn\ and \gat. 
We observe that even when the base architecture changes, both \ours[N] and \ours[E] achieve competitive accuracy compared to the respective \vanilla\ approach that simply trains the GNN architecture using all the information from the original graph. We refer the reader to \cref{app:comparison} for the complete results.

\subsection{Performance comparison}

In this section, we conduct a comprehensive evaluation of \ours\ by comparing it with baselines and competing methods across nine datasets. 

We assess the performance on the test set using three key metrics: accuracy (the larger the better $\uparrow$), as well as the percentage of nodes (for \ours[N]) or edges (for \ours[E]) kept in the subgraph (the lower the better $\downarrow$). 
Decreasing these percentages increases graph sparsity, enhancing model interpretability.

\begin{table}[t]
	\caption{\textbf{Model comparison results.} We show the mean over 5 independent runs and the standard deviation as the subindex. All metrics are shown in percentage. The last rows include the average ranking of the model across datasets. Best performing models on average are indicated in bold.}
	\centering
	\begin{tabular}{lcccc|cccc}
		\toprule
		\multirow{2}{*}{\textbf{Dataset}} & \multirow{2}{*}{\textbf{Metric}} & \multicolumn{3}{c}{\textbf{\emph{Full} Models}} & \multicolumn{4}{c}{\textbf{\emph{Sparse} Models}} \\ 
		\cmidrule(lr){3-5} \cmidrule(lr){6-9}
		& & \textbf{\vanilla} & \textbf{SAGPool} & \textbf{DiffPool} & \textbf{SUGAR} & \textbf{GPool} & \textbf{CORES$_N$} & \textbf{CORES$_E$}\\ 
		\midrule
		\multirow{3}{*}{BZR} & \textbf{Accuracy} & 82.44$_{5.00}$ & 85.85$_{2.67}$ & 70.24$_{5.95}$ & - & 80.98$_{4.01}$ & 84.88$_{4.01}$ & 81.95$_{5.62}$\\ 
		& \textbf{Node Ratio} & - & - & - & - & 96.46$_{0.19}$ & 15.91$_{5.30}$ & 100.00$_{0.00}$\\ 
		& \textbf{Edge Ratio} & - & - & - & - & 96.40$_{0.45}$ & 10.84$_{5.80}$ & 76.03$_{3.72}$\\ 
		\cmidrule(lr){2-9}
		\multirow{3}{*}{COX2} & \textbf{Accuracy} & 82.13$_{5.12}$ & 80.43$_{3.16}$ & 65.83$_{6.65}$ & - & 80.43$_{3.50}$ & 82.98$_{3.36}$ & 85.11$_{3.01}$\\ 
		& \textbf{Node Ratio} & - & - & - & - & 91.15$_{0.14}$ & 86.44$_{3.48}$ & 100.00$_{0.00}$\\ 
		& \textbf{Edge Ratio} & - & - & - & - & 89.20$_{1.08}$ & 77.63$_{6.60}$ & 58.30$_{28.00}$\\ 
		\cmidrule(lr){2-9}
		\multirow{3}{*}{DD} & \textbf{Accuracy} & 78.81$_{3.44}$ & 74.07$_{4.09}$ & - & - & 77.46$_{3.31}$ & 75.42$_{5.22}$ & 75.25$_{2.77}$\\ 
		& \textbf{Node Ratio} & - & - & - & - & 40.23$_{0.03}$ & 47.53$_{15.04}$ & 100.00$_{0.00}$\\ 
		& \textbf{Edge Ratio} & - & - & - & - & 15.41$_{0.26}$ & 25.88$_{14.79}$ & 51.55$_{2.42}$\\ 
		\cmidrule(lr){2-9}
		\multirow{3}{*}{ENZYMES} & \textbf{Accuracy} & 61.67$_{8.50}$ & 45.90$_{5.18}$ & 31.00$_{6.66}$ & 16.67$_{23.57}$ & 62.62$_{5.36}$ & 62.33$_{6.08}$ & 64.92$_{3.40}$\\ 
		& \textbf{Node Ratio} & - & - & - & - & 96.67$_{0.14}$ & 80.67$_{1.16}$ & 100.00$_{0.00}$\\ 
		& \textbf{Edge Ratio} & - & - & - & - & 93.37$_{0.14}$ & 66.90$_{1.66}$ & 72.71$_{1.00}$\\ 
		\cmidrule(lr){2-9}
		\multirow{3}{*}{MUTAG} & \textbf{Accuracy} & 85.71$_{8.75}$ & 78.57$_{5.05}$ & 69.00$_{12.21}$ & 76.34$_{3.80}$ & 81.43$_{3.91}$ & 82.86$_{6.39}$ & 82.14$_{5.98}$\\ 
		& \textbf{Node Ratio} & - & - & - & - & 51.38$_{0.12}$ & 52.38$_{7.67}$ & 100.00$_{0.00}$\\ 
		& \textbf{Edge Ratio} & - & - & - & - & 36.56$_{9.20}$ & 36.31$_{11.32}$ & 61.09$_{12.39}$\\ 
		\cmidrule(lr){2-9}
		\multirow{3}{*}{NCI1} & \textbf{Accuracy} & 76.89$_{2.03}$ & 69.29$_{2.12}$ & 69.93$_{3.24}$ & 49.95$_{35.58}$ & 77.77$_{1.93}$ & 73.34$_{6.30}$ & 74.34$_{7.53}$\\ 
		& \textbf{Node Ratio} & - & - & - & - & 96.89$_{0.08}$ & 89.13$_{13.88}$ & 100.00$_{0.00}$\\ 
		& \textbf{Edge Ratio} & - & - & - & - & 94.47$_{0.28}$ & 86.61$_{20.20}$ & 84.06$_{27.86}$\\ 
		\cmidrule(lr){2-9}
		\multirow{3}{*}{NCI109} & \textbf{Accuracy} & 78.64$_{2.69}$ & 74.14$_{0.85}$ & 67.49$_{2.60}$ & 49.65$_{35.71}$ & 75.45$_{3.30}$ & 73.03$_{1.90}$ & 75.74$_{2.04}$\\ 
		& \textbf{Node Ratio} & - & - & - & - & 96.79$_{0.04}$ & 66.31$_{5.69}$ & 100.00$_{0.00}$\\ 
		& \textbf{Edge Ratio} & - & - & - & - & 93.65$_{0.22}$ & 61.26$_{7.22}$ & 71.34$_{7.62}$\\ 
		\cmidrule(lr){2-9}
		\multirow{3}{*}{PROTEINS} & \textbf{Accuracy} & 75.18$_{2.04}$ & 70.54$_{3.63}$ & 66.42$_{7.19}$ & 59.57$_{43.01}$ & 72.68$_{1.85}$ & 73.75$_{4.02}$ & 74.11$_{2.82}$\\ 
		& \textbf{Node Ratio} & - & - & - & - & 41.92$_{0.30}$ & 87.01$_{20.04}$ & 100.00$_{0.00}$\\ 
		& \textbf{Edge Ratio} & - & - & - & - & 19.42$_{0.64}$ & 78.55$_{32.05}$ & 83.31$_{37.32}$\\ 
		\cmidrule(lr){2-9}
		\multirow{3}{*}{PTC} & \textbf{Accuracy} & 63.43$_{1.28}$ & 61.14$_{8.23}$ & 54.44$_{11.10}$ & 56.14$_{8.95}$ & 59.43$_{3.73}$ & 64.00$_{3.26}$ & 63.43$_{1.28}$\\ 
		& \textbf{Node Ratio} & - & - & - & - & 43.52$_{0.67}$ & 96.02$_{8.80}$ & 100.00$_{0.00}$\\ 
		& \textbf{Edge Ratio} & - & - & - & - & 24.80$_{3.48}$ & 93.19$_{15.13}$ & 88.61$_{15.64}$\\ 
		\bottomrule
		\multirow{3}{*}{\textbf{Avg. Rank}} & \textbf{Accuracy} & 2.11 & 4.44 & 6.12 & 6.67 & 3.44 & 2.67 & 2.67\\ 
		\cmidrule(lr){2-9}
		& \textbf{Node Ratio} & - & - & - & - & 1.56 & 1.44 & - \\ 
		\cmidrule(lr){2-9}
		& \textbf{Edge Ratio} & - & - & - & - & 2.22 & 1.67 & 2.11\\ 
		\bottomrule
	\end{tabular}
	\label{tab:comparison-gin}
\end{table}

\myparagraph{Baselines}
We divided the baselines into two categories: \emph{Full} model (use complete graph information) and \emph{Sparse} (use part of the graph information) model approaches. 
In the \emph{Full} model category, we compare with the following approaches:
\vanilla\ approach representing training the  base GNN architecture without any sparsity consideration,
\diffpool\ \cite{diffpool}, and 
\topksoft\ \cite{Lee2019SelfAttentionGP, Knyazev2019UnderstandingAA}---which uses node embeddings coming from a GNN to select the top $k$ nodes, thus incorporating global information.
For the \emph{Sparse} models, we compare with:
\sugar\ \cite{SUGAR} \footnote{We use the official implementation of \sugar\ to extract the results on the above datasets. We believe our evaluation of a held-out test set may explain the disparity between the results we report and those in \cite{SUGAR}. Instead, the available implementation of \sugar\ evaluates on the same validation set used to select the best model. } (for the datasets that their code can run) and
\topkhard\ \cite{Gao2019GraphU,Cangea2018TowardsSH, Knyazev2019UnderstandingAA}--- where the node embeddings used to select the top $k$ come from a multi-layer perceptron, hence the subgraph does not contain global information.
Here, we present the results using \gin\ \cite{GIN} as the base GNN architecture. Refer to \cref{app:comparison} for the results using \gcn\ and \gat.

\myparagraph{Quantitative results}
\cref{tab:comparison-gin} summarizes the results. 
In terms of accuracy, we observe that while \vanilla\ consistently achieves the highest accuracy on most datasets, \ours[E] and \ours[N] secure the second and third-best positions, outperforming all other \emph{Full} models and all \emph{Sparse} models, which highlights the competitive performance of the proposed approach.
On the other hand, \sugar\ consistently underperforms across all datasets, and its results exhibit high variance.
Regarding interpretability (understood as sparsity), we perform a node and edge ratio analysis on the three best-performing \emph{Sparse} models in terms of accuracy: \ours[N], \ours[E], and \topkhard.
It is important to note that \ours[E] consistently maintains a node ratio of 100\%, due to its edge removal strategy that preserves all nodes.
Among these models, \ours[N] achieves the highest rankings in both node and edge ratios, indicating its effectiveness in removing a larger number of nodes from the original graph. 
Consequently, the graph classifier operates on a reduced, and thus more interpretable, subgraph.
Finally, we find it interesting to analyze the results for the \ptc\ dataset (bottom row). We observe that  \ours, in both removal modes, achieves the worst sparsity results but at the same time the best performance among all models.
In terms of sparsity, \ours[N] keeps 96.02\% of the nodes, while \ours[E] keeps 88.61\% of the edges. 
In terms of accuracy, all the models achieve less than  65\%, which is notably low for a binary classification problem. 
This indicates that the classifier makes a substantial number of errors. 
This observation is important considering the design of the proposed reward function of \ours\ (see \cref{eq:reward}). 
For this dataset, the reward function places \ours\ in a scenario where it seeks to enhance performance, corresponding to the middle case, or penalize sparsity, corresponding to the bottom case of the reward function.
In essence, \ours\ behaves in accordance with the reward function.

\begin{figure}
	\centering
	
	\includegraphics[width=0.5\textwidth]{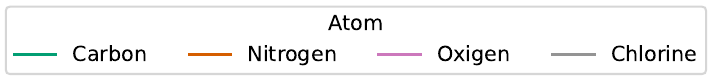}
	\includegraphics[width=0.9\textwidth]{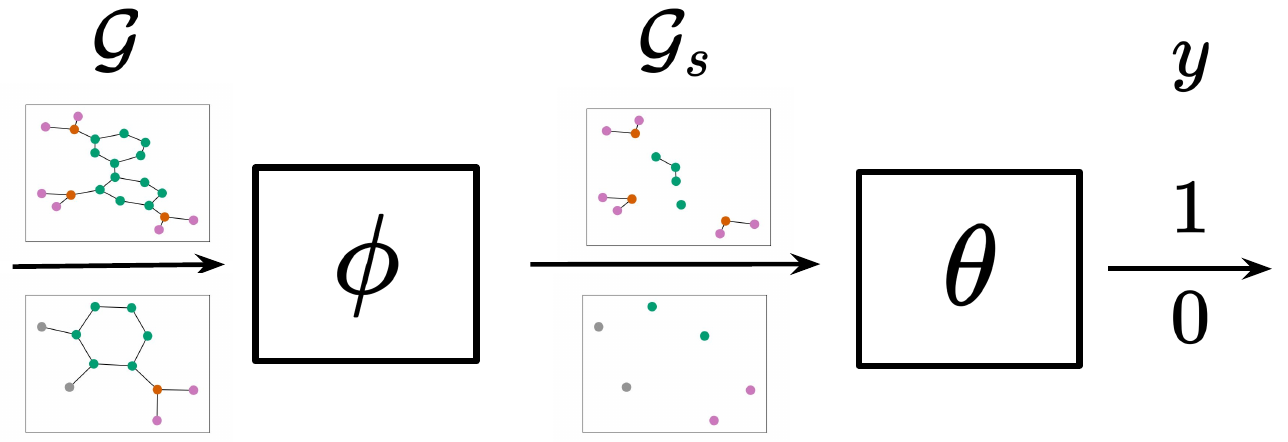}
	\captionsetup{font=small}
	\caption{Illustration of the predictive subgraph obtained by \ours[N] for a mutagenic (top) and non-mutagenic (bottom) graph of the \mutag\ dataset.}
	\label{fig:pipeline_mutag}
\end{figure}

\myparagraph{Qualitative results}
\cref{fig:pipeline_mutag} presents two examples of the predictive subgraphs discovered by \ours[N] for the \mutag\ dataset, showcasing positive (top) and negative (bottom) instances. 
In this dataset, the positive class ($y=1$) represents mutagenic molecules, while the negative class ($y=0$) corresponds to non-mutagenic molecules. The left side of the figure shows the original graphs, while the center illustrates the predictive subgraphs identified by \ours[N].
We observe that for the mutagenic graph (top), \ours[N] keeps $N0_2$ nitrogen dioxide compounds and some carbon atoms, while for the non-mutagenic graph (bottom), it captures chlorine carbon atoms as relevant for the classification task. 
These findings align with prior in the chemical domain \cite{Debnath1991StructureactivityRO}.

\myparagraph{Time complexity} 
Training and inference are significantly faster in methods that employ a straightforward forward pass compared to RL-based methods like \sugar\ and \ours. Notably, our findings indicate that \ours\ is at least as fast as \sugar\ during training and can be up to 10 times faster during inference. For comprehensive quantitative results across all datasets, please refer to \cref{app:time-complexity}.

\section{Conclusions}

Graph Neural Networks (GNNs) have demonstrated exceptional performance in graph-level tasks,  but suffer from interpretability issues due to their complexity.
Existing interpretability research in GNNs mainly focuses on post-hoc explanations. 
Approaches that aim to introduce \emph{sparsity} during training often provide predictive subgraphs that are not \emph{faithful} with the classifier predictions. 
Also, these methods often rely on complete graph information and/or make strong assumptions about the sparser graphs.
In this work, we introduced \ours, a novel GNN training approach that simultaneously identifies predictive subgraphs by removing edges and/or nodes, without imposing assumptions about subgraph structures; and optimizes the performance of the graph classification task. 
The optimization for performance refers to the conventional supervised problem paradigm. 
We then leverage reinforcement learning to identify predictive subgraphs, using a policy that allows both edge and node removal modes.
One key aspect of \ours\ is the design of the reward function. This function allows the practitioner to introduce inductive biases towards either sparse or high-performing predictive subgraphs and it incorporates conformal predictions \cite{Angelopoulos2021AGI} to account for classifier uncertainty.
Our empirical evaluation, conducted on nine graph classification datasets, provides evidence that our approach not only competes in performance with the baselines that use the complete graph information but also relies on significantly sparser subgraphs. 
Consequently, the resulting GNN-based predictions are more interpretable, addressing the primary motivation behind our work.

\myparagraph{Practical limitations}
The main limitation of our proposed approach lies in the significantly increased training and inference time, introduced by the reinforcement learning component,  compared to the baselines that do not use reinforcement learning.

\myparagraph{Future work}
Future research could focus on improving the efficiency of the reinforcement learning component to accelerate the training process of \ours.
Another interesting avenue for future work could involve modifying the current reward function to penalize specific types of errors or undesirable structures, such as isolated nodes.
Additionally, it could also be interesting to explore the application of \ours\ to graph regression tasks and \ours[E] to node classification tasks.

 \section*{Acknowledgements}

 Pablo S\'anchez Mart\'in thanks the German Research Foundation through the Cluster of Excellence “Machine Learning – New Perspectives for Science”, EXC 2064/1, project number 390727645 for generous funding support. 
 The authors thank the International Max Planck Research School for Intelligent Systems (IMPRS-IS) for supporting Pablo S\'anchez Mart\'in.

	\bibliographystyle{plainnat} %
	\bibliography{references.bib}

	\newpage
	\appendix
	\section{Training details}
\label{app:training}

In this section, we provide further details of the experimental setup used to obtain our results.
For all experiments, we train the models using a fixed seed of 0, and the reported results represent the mean and standard deviation over five different dataset splits.
We run the models up to  1000 epochs, doing early stopping if the accuracy does not improve for 500 epochs. All experiments were conducted on a single CPU with 8GB of RAM.
We carry out hyperparameter tuning using random sampling. We explored 20 different configurations for each model, data, and GNN architecture. Complete details regarding the cross-validated hyperparameters can be found in our GitHub repository at \url{https://github.com/psanch21/CORES}.
\cref{tab:params-gnn} presents the best configurations achieved for each dataset and architecture for the \vanilla\ GNN baselines. These configurations were subsequently used for training \ours. The best configuration for \ours[N] is detailed in \cref{tab:params-cores-node}, while for \ours[E], it can be found in \cref{tab:params-cores-edge}.
The best configurations for \topksoft\ and \topkhard\ are outlined in \cref{tab:params-topksoft} and \cref{tab:params-topkhard}, respectively.

\begin{table}[t]
\caption{\textbf{Best configuration for the \vanilla\ GNN.} Split sizes ($\#$0), Batch size ($\#$1), Dropout rate ($\#$2), Batch normalizing ($\#$3), Dimension of hidden layers ($\#$4), Number of GNN layers ($\#$5), Global pooling type ($\#$6), Classifier scheduler factor ($\#$7), Classifier learning rate ($\#$8), $\epsilon$ ($\#$9), Trainable $\epsilon$ ($\#$10), Number of heads ($\#$11).}
\centering
{\tiny
\begin{tabular}{lcccccccccc}
\toprule
\multirow{2}{*}{\textbf{Param}} & \multirow{2}{*}{\textbf{Archi}} & \multicolumn{9}{c}{\textbf{Datasets}} \\ 
\cmidrule(lr){3-11}
& & \textbf{BZR} & \textbf{COX2} & \textbf{DD} & \textbf{ENZYMES} & \textbf{MUTAG} & \textbf{NCI1} & \textbf{NCI109} & \textbf{PROTEINS} & \textbf{PTC}\\ 
\midrule
\multirow{3}{*}{ $\#$0} & \textbf{GIN} & [0.6, 0.3, 0.1] & [0.4, 0.5, 0.1] & [0.6, 0.3, 0.1] & [0.5, 0.4, 0.1] & [0.4, 0.5, 0.1] & [0.5, 0.4, 0.1] & [0.4, 0.5, 0.1] & [0.4, 0.5, 0.1] & [0.6, 0.3, 0.1]\\ 
 & \textbf{GAT} & [0.6, 0.3, 0.1] & [0.5, 0.4, 0.1] & [0.5, 0.4, 0.1] & [0.5, 0.4, 0.1] & [0.5, 0.4, 0.1] & [0.6, 0.3, 0.1] & [0.6, 0.3, 0.1] & [0.4, 0.5, 0.1] & [0.5, 0.4, 0.1]\\ 
 & \textbf{GCN} & [0.6, 0.3, 0.1] & [0.5, 0.4, 0.1] & [0.6, 0.3, 0.1] & [0.5, 0.4, 0.1] & [0.5, 0.4, 0.1] & [0.6, 0.3, 0.1] & [0.6, 0.3, 0.1] & [0.4, 0.5, 0.1] & [0.4, 0.5, 0.1]\\ 
\cmidrule(lr){2-11}
\multirow{3}{*}{ $\#$1} & \textbf{GIN} & 16 & 16 & 32 & 16 & 16 & 16 & 32 & 16 & 32\\ 
 & \textbf{GAT} & 32 & 16 & 32 & 16 & 16 & 32 & 32 & 16 & 16\\ 
 & \textbf{GCN} & 16 & 16 & 32 & 16 & 32 & 32 & 16 & 32 & 32\\ 
\cmidrule(lr){2-11}
\multirow{3}{*}{ $\#$2} & \textbf{GIN} & 0.0 & 0.2 & 0.5 & 0.2 & 0.0 & 0.0 & 0.1 & 0.1 & 0.4\\ 
 & \textbf{GAT} & 0.1 & 0.2 & 0.3 & 0.4 & 0.3 & 0.2 & 0.1 & 0.1 & 0.5\\ 
 & \textbf{GCN} & 0.3 & 0.1 & 0.3 & 0.0 & 0.4 & 0.0 & 0.1 & 0.4 & 0.5\\ 
\cmidrule(lr){2-11}
\multirow{3}{*}{ $\#$3} & \textbf{GIN} & True & False & True & True & True & True & True & False & False\\ 
 & \textbf{GAT} & True & True & False & True & True & True & False & False & True\\ 
 & \textbf{GCN} & True & False & False & True & False & True & True & False & False\\ 
\cmidrule(lr){2-11}
\multirow{3}{*}{ $\#$4} & \textbf{GIN} & 128 & 64 & 32 & 64 & 16 & 128 & 128 & 128 & 16\\ 
 & \textbf{GAT} & 16 & 128 & 16 & 128 & 32 & 128 & 128 & 16 & 16\\ 
 & \textbf{GCN} & 64 & 128 & 32 & 128 & 16 & 64 & 128 & 32 & 32\\ 
\cmidrule(lr){2-11}
\multirow{3}{*}{ $\#$5} & \textbf{GIN} & 3 & 1 & 1 & 1 & 3 & 1 & 3 & 1 & 3\\ 
 & \textbf{GAT} & 1 & 2 & 1 & 4 & 4 & 4 & 3 & 2 & 4\\ 
 & \textbf{GCN} & 1 & 4 & 3 & 3 & 4 & 3 & 4 & 1 & 1\\ 
\cmidrule(lr){2-11}
\multirow{3}{*}{ $\#$6} & \textbf{GIN} & ['mean', 'add'] & ['add'] & ['mean', 'add'] & ['mean'] & ['mean', 'add'] & ['mean'] & ['mean'] & ['add'] & ['mean']\\ 
 & \textbf{GAT} & ['add'] & ['mean'] & ['mean', 'add'] & ['mean'] & ['mean', 'add'] & ['add'] & ['mean', 'add'] & ['add'] & ['mean', 'add']\\ 
 & \textbf{GCN} & ['mean'] & ['mean', 'add'] & ['mean', 'add'] & ['mean'] & ['mean', 'add'] & ['mean', 'add'] & ['add'] & ['add'] & ['mean']\\ 
\cmidrule(lr){2-11}
\multirow{3}{*}{ $\#$7} & \textbf{GIN} & 0.95 & 0.99 & 0.99 & 0.9 & 0.95 & 0.9 & 0.99 & 0.9 & 0.99\\ 
 & \textbf{GAT} & 0.95 & 0.95 & 0.99 & 0.95 & 0.99 & 0.95 & 0.99 & 0.95 & 0.9\\ 
 & \textbf{GCN} & 0.99 & 0.99 & 0.99 & 0.95 & 0.99 & 0.99 & 0.9 & 0.99 & 0.99\\ 
\cmidrule(lr){2-11}
\multirow{3}{*}{ $\#$8} & \textbf{GIN} & 0.0001 & 0.001 & 0.005 & 0.0001 & 0.001 & 0.01 & 0.001 & 0.0005 & 0.005\\ 
 & \textbf{GAT} & 0.01 & 0.0001 & 0.0001 & 0.001 & 0.005 & 0.005 & 0.001 & 0.005 & 0.01\\ 
 & \textbf{GCN} & 0.01 & 0.0001 & 0.001 & 0.01 & 0.005 & 0.005 & 0.001 & 0.001 & 0.01\\ 
\cmidrule(lr){2-11}
\multirow{3}{*}{ $\#$9} & \textbf{GIN} & 0.3 & 0.2 & 0.3 & 0.0 & 0.2 & 0.2 & 0.2 & 0.0 & 0.2\\ 
 & \textbf{GAT} & - & - & - & - & - & - & - & - & -\\ 
 & \textbf{GCN} & - & - & - & - & - & - & - & - & -\\ 
\cmidrule(lr){2-11}
\multirow{3}{*}{ $\#$10} & \textbf{GIN} & False & True & True & True & True & False & True & False & False\\ 
 & \textbf{GAT} & - & - & - & - & - & - & - & - & -\\ 
 & \textbf{GCN} & - & - & - & - & - & - & - & - & -\\ 
\cmidrule(lr){2-11}
\multirow{3}{*}{ $\#$11} & \textbf{GIN} & - & - & - & - & - & - & - & - & -\\ 
 & \textbf{GAT} & 2.0 & 4.0 & 4.0 & 2.0 & 2.0 & 4.0 & 4.0 & 2.0 & 4.0\\ 
 & \textbf{GCN} & - & - & - & - & - & - & - & - & -\\ 
\bottomrule
\end{tabular}
}
\label{tab:params-gnn}
\end{table}

\begin{table}[t]
\caption{\textbf{Best configuration for \topksoft.} Split sizes ($\#$0), Batch size ($\#$1), Early stopping clf. patience ($\#$2), Dropout rate ($\#$3), Batch normalizing ($\#$4), Dimension of hidden layers ($\#$5), Number of GNN layers ($\#$6), Global pooling type ($\#$7), Classifier scheduler factor ($\#$8), Classifier learning rate ($\#$9), TopK multiplier ($\#$10), TopK ratio ($\#$11), Number of heads ($\#$12), $\epsilon$ ($\#$13), Trainble $\epsilon$ ($\#$14)}
\centering
{\tiny
\begin{tabular}{lcccccccccc}
\toprule
\multirow{2}{*}{\textbf{Param}} & \multirow{2}{*}{\textbf{Archi}} & \multicolumn{9}{c}{\textbf{Datasets}} \\ 
\cmidrule(lr){3-11}
& & \textbf{BZR} & \textbf{COX2} & \textbf{DD} & \textbf{ENZYMES} & \textbf{MUTAG} & \textbf{NCI1} & \textbf{NCI109} & \textbf{PROTEINS} & \textbf{PTC}\\ 
\midrule
\multirow{3}{*}{ $\#$0} & \textbf{GCN} & [0.5, 0.4, 0.1] & [0.5, 0.4, 0.1] & [0.4, 0.5, 0.1] & [0.5, 0.4, 0.1] & [0.5, 0.4, 0.1] & [0.5, 0.4, 0.1] & [0.5, 0.4, 0.1] & [0.4, 0.5, 0.1] & [0.6, 0.3, 0.1]\\ 
 & \textbf{GAT} & [0.4, 0.5, 0.1] & [0.4, 0.5, 0.1] & [0.4, 0.5, 0.1] & [0.5, 0.4, 0.1] & [0.4, 0.5, 0.1] & [0.6, 0.3, 0.1] & [0.5, 0.4, 0.1] & [0.4, 0.5, 0.1] & [0.4, 0.5, 0.1]\\ 
 & \textbf{GIN} & [0.5, 0.4, 0.1] & [0.5, 0.4, 0.1] & [0.5, 0.4, 0.1] & [0.6, 0.3, 0.1] & [0.5, 0.4, 0.1] & [0.5, 0.4, 0.1] & [0.6, 0.3, 0.1] & [0.6, 0.3, 0.1] & [0.5, 0.4, 0.1]\\ 
\cmidrule(lr){2-11}
\multirow{3}{*}{ $\#$1} & \textbf{GCN} & 16 & 16 & 16 & 16 & 32 & 16 & 16 & 16 & 32\\ 
 & \textbf{GAT} & 16 & 16 & 16 & 16 & 16 & 16 & 32 & 16 & 16\\ 
 & \textbf{GIN} & 16 & 16 & 16 & 32 & 32 & 16 & 32 & 32 & 16\\ 
\cmidrule(lr){2-11}
\multirow{3}{*}{ $\#$2} & \textbf{GCN} & 800 & 650 & 500 & 650 & 500 & 650 & 650 & 500 & 650\\ 
 & \textbf{GAT} & 800 & 800 & 500 & 800 & 800 & 650 & 500 & 800 & 800\\ 
 & \textbf{GIN} & 800 & 500 & 650 & 500 & 800 & 800 & 500 & 500 & 650\\ 
\cmidrule(lr){2-11}
\multirow{3}{*}{ $\#$3} & \textbf{GCN} & 0.4 & 0.0 & 0.5 & 0.1 & 0.4 & 0.0 & 0.1 & 0.5 & 0.4\\ 
 & \textbf{GAT} & 0.3 & 0.3 & 0.1 & 0.1 & 0.4 & 0.1 & 0.4 & 0.3 & 0.3\\ 
 & \textbf{GIN} & 0.2 & 0.3 & 0.5 & 0.0 & 0.4 & 0.2 & 0.0 & 0.0 & 0.5\\ 
\cmidrule(lr){2-11}
\multirow{3}{*}{ $\#$4} & \textbf{GCN} & True & False & False & True & True & False & True & False & False\\ 
 & \textbf{GAT} & False & False & False & False & True & True & False & False & False\\ 
 & \textbf{GIN} & True & False & False & True & False & True & True & True & False\\ 
\cmidrule(lr){2-11}
\multirow{3}{*}{ $\#$5} & \textbf{GCN} & 128 & 32 & 32 & 128 & 32 & 32 & 128 & 32 & 64\\ 
 & \textbf{GAT} & 16 & 16 & 64 & 128 & 16 & 32 & 16 & 16 & 16\\ 
 & \textbf{GIN} & 32 & 128 & 128 & 64 & 16 & 32 & 64 & 64 & 128\\ 
\cmidrule(lr){2-11}
\multirow{3}{*}{ $\#$6} & \textbf{GCN} & 4 & 1 & 4 & 4 & 1 & 1 & 4 & 4 & 3\\ 
 & \textbf{GAT} & 1 & 1 & 2 & 2 & 2 & 3 & 2 & 1 & 1\\ 
 & \textbf{GIN} & 4 & 4 & 3 & 1 & 2 & 4 & 1 & 1 & 3\\ 
\cmidrule(lr){2-11}
\multirow{3}{*}{ $\#$7} & \textbf{GCN} & ['add'] & ['mean', 'add'] & ['add'] & ['mean'] & ['mean', 'add'] & ['mean', 'add'] & ['mean'] & ['add'] & ['mean']\\ 
 & \textbf{GAT} & ['mean', 'add'] & ['mean', 'add'] & ['mean', 'add'] & ['mean', 'add'] & ['mean', 'add'] & ['mean'] & ['mean', 'add'] & ['mean', 'add'] & ['mean', 'add']\\ 
 & \textbf{GIN} & ['mean'] & ['mean'] & ['mean'] & ['add'] & ['mean', 'add'] & ['mean'] & ['add'] & ['add'] & ['mean']\\ 
\cmidrule(lr){2-11}
\multirow{3}{*}{ $\#$8} & \textbf{GCN} & 0.99 & 0.95 & 0.99 & 0.95 & 0.9 & 0.95 & 0.95 & 0.99 & 0.9\\ 
 & \textbf{GAT} & 0.9 & 0.9 & 0.9 & 0.99 & 0.99 & 0.95 & 0.99 & 0.9 & 0.9\\ 
 & \textbf{GIN} & 0.9 & 0.99 & 0.95 & 0.95 & 0.99 & 0.9 & 0.95 & 0.95 & 0.95\\ 
\cmidrule(lr){2-11}
\multirow{3}{*}{ $\#$9} & \textbf{GCN} & 0.01 & 0.005 & 0.0005 & 0.0001 & 0.01 & 0.005 & 0.0001 & 0.0005 & 0.0001\\ 
 & \textbf{GAT} & 0.01 & 0.01 & 0.01 & 0.005 & 0.005 & 0.0001 & 0.005 & 0.01 & 0.01\\ 
 & \textbf{GIN} & 0.001 & 0.0005 & 0.0005 & 0.001 & 0.005 & 0.001 & 0.001 & 0.001 & 0.0005\\ 
\cmidrule(lr){2-11}
\multirow{3}{*}{ $\#$10} & \textbf{GCN} & 0.5 & 2.0 & 0.5 & 1.5 & 1.0 & 2.0 & 1.5 & 0.5 & 0.5\\ 
 & \textbf{GAT} & 1.0 & 1.0 & 2.0 & 1.0 & 1.5 & 1.5 & 0.5 & 1.0 & 1.0\\ 
 & \textbf{GIN} & 1.5 & 2.0 & 0.5 & 0.5 & 0.5 & 1.5 & 0.5 & 0.5 & 0.5\\ 
\cmidrule(lr){2-11}
\multirow{3}{*}{ $\#$11} & \textbf{GCN} & 0.95 & 0.95 & 0.9 & 0.3 & 0.3 & 0.95 & 0.3 & 0.9 & 0.2\\ 
 & \textbf{GAT} & 0.9 & 0.9 & 0.5 & 0.9 & 0.2 & 0.6 & 0.8 & 0.9 & 0.9\\ 
 & \textbf{GIN} & 0.3 & 0.4 & 0.8 & 0.3 & 0.8 & 0.3 & 0.3 & 0.3 & 0.8\\ 
\cmidrule(lr){2-11}
\multirow{3}{*}{ $\#$12} & \textbf{GCN} & - & - & - & - & - & - & - & - & -\\ 
 & \textbf{GAT} & 1.0 & 1.0 & 1.0 & 1.0 & 1.0 & 1.0 & 4.0 & 1.0 & 1.0\\ 
 & \textbf{GIN} & - & - & - & - & - & - & - & - & -\\ 
\cmidrule(lr){2-11}
\multirow{3}{*}{ $\#$13} & \textbf{GCN} & - & - & - & - & - & - & - & - & -\\ 
 & \textbf{GAT} & - & - & - & - & - & - & - & - & -\\ 
 & \textbf{GIN} & 0.4 & 0.2 & 0.4 & 0.4 & 0.3 & 0.4 & 0.4 & 0.4 & 0.4\\ 
\cmidrule(lr){2-11}
\multirow{3}{*}{ $\#$14} & \textbf{GCN} & - & - & - & - & - & - & - & - & -\\ 
 & \textbf{GAT} & - & - & - & - & - & - & - & - & -\\ 
 & \textbf{GIN} & False & True & True & True & False & False & True & True & True\\ 
\bottomrule
\end{tabular}
}
\label{tab:params-topksoft}
\end{table}

\begin{table}[t]
\caption{\textbf{Best configuration for the \topkhard.} Split sizes ($\#$0), Batch size ($\#$1), Early stopping clf. patience ($\#$2), Dropout rate ($\#$3), Batch normalizing ($\#$4), Dimension of hidden layers ($\#$5), Number of GNN layers ($\#$6), Global pooling type ($\#$7), Classifier scheduler factor ($\#$8), Classifier learning rate ($\#$9), TopK multiplier ($\#$10), TopK ratio ($\#$11), Number of heads ($\#$12), $\epsilon$ ($\#$13), Trainble $\epsilon$ ($\#$14).}
\centering
{\tiny
\begin{tabular}{lcccccccccc}
\toprule
\multirow{2}{*}{\textbf{Param}} & \multirow{2}{*}{\textbf{Archi}} & \multicolumn{9}{c}{\textbf{Datasets}} \\ 
\cmidrule(lr){3-11}
& & \textbf{BZR} & \textbf{COX2} & \textbf{DD} & \textbf{ENZYMES} & \textbf{MUTAG} & \textbf{NCI1} & \textbf{NCI109} & \textbf{PROTEINS} & \textbf{PTC}\\ 
\midrule
\multirow{3}{*}{ $\#$0} & \textbf{GAT} & [0.5, 0.4, 0.1] & [0.6, 0.3, 0.1] & [0.6, 0.3, 0.1] & [0.4, 0.5, 0.1] & [0.4, 0.5, 0.1] & [0.4, 0.5, 0.1] & [0.4, 0.5, 0.1] & [0.4, 0.5, 0.1] & [0.4, 0.5, 0.1]\\ 
 & \textbf{GIN} & [0.4, 0.5, 0.1] & [0.4, 0.5, 0.1] & [0.4, 0.5, 0.1] & [0.4, 0.5, 0.1] & [0.5, 0.4, 0.1] & [0.4, 0.5, 0.1] & [0.4, 0.5, 0.1] & [0.4, 0.5, 0.1] & [0.5, 0.4, 0.1]\\ 
 & \textbf{GCN} & [0.4, 0.5, 0.1] & [0.4, 0.5, 0.1] & [0.4, 0.5, 0.1] & [0.4, 0.5, 0.1] & [0.5, 0.4, 0.1] & [0.4, 0.5, 0.1] & [0.4, 0.5, 0.1] & [0.4, 0.5, 0.1] & [0.6, 0.3, 0.1]\\ 
\cmidrule(lr){2-11}
\multirow{3}{*}{ $\#$1} & \textbf{GAT} & 16 & 16 & 16 & 32 & 32 & 32 & 32 & 32 & 32\\ 
 & \textbf{GIN} & 32 & 16 & 16 & 32 & 16 & 32 & 32 & 16 & 32\\ 
 & \textbf{GCN} & 16 & 16 & 32 & 32 & 16 & 32 & 32 & 32 & 16\\ 
\cmidrule(lr){2-11}
\multirow{3}{*}{ $\#$2} & \textbf{GAT} & 500 & 650 & 800 & 500 & 500 & 500 & 500 & 650 & 800\\ 
 & \textbf{GIN} & 500 & 800 & 800 & 500 & 800 & 500 & 500 & 800 & 800\\ 
 & \textbf{GCN} & 650 & 650 & 500 & 500 & 650 & 500 & 500 & 800 & 650\\ 
\cmidrule(lr){2-11}
\multirow{3}{*}{ $\#$3} & \textbf{GAT} & 0.0 & 0.3 & 0.5 & 0.1 & 0.1 & 0.1 & 0.1 & 0.1 & 0.3\\ 
 & \textbf{GIN} & 0.1 & 0.0 & 0.5 & 0.1 & 0.4 & 0.1 & 0.1 & 0.5 & 0.2\\ 
 & \textbf{GCN} & 0.0 & 0.0 & 0.4 & 0.1 & 0.0 & 0.3 & 0.1 & 0.5 & 0.1\\ 
\cmidrule(lr){2-11}
\multirow{3}{*}{ $\#$4} & \textbf{GAT} & True & True & False & True & True & True & True & False & False\\ 
 & \textbf{GIN} & True & True & False & True & False & True & True & False & False\\ 
 & \textbf{GCN} & True & True & False & True & True & True & True & True & False\\ 
\cmidrule(lr){2-11}
\multirow{3}{*}{ $\#$5} & \textbf{GAT} & 16 & 64 & 16 & 64 & 64 & 64 & 64 & 16 & 64\\ 
 & \textbf{GIN} & 64 & 32 & 16 & 64 & 32 & 64 & 64 & 16 & 128\\ 
 & \textbf{GCN} & 128 & 128 & 128 & 64 & 32 & 128 & 64 & 16 & 16\\ 
\cmidrule(lr){2-11}
\multirow{3}{*}{ $\#$6} & \textbf{GAT} & 2 & 4 & 1 & 2 & 2 & 2 & 2 & 1 & 3\\ 
 & \textbf{GIN} & 2 & 1 & 1 & 2 & 1 & 2 & 2 & 1 & 4\\ 
 & \textbf{GCN} & 2 & 2 & 2 & 2 & 4 & 3 & 2 & 3 & 3\\ 
\cmidrule(lr){2-11}
\multirow{3}{*}{ $\#$7} & \textbf{GAT} & ['mean', 'add'] & ['mean'] & ['mean', 'add'] & ['mean'] & ['mean'] & ['mean'] & ['mean'] & ['add'] & ['add']\\ 
 & \textbf{GIN} & ['mean'] & ['mean'] & ['mean', 'add'] & ['mean'] & ['mean', 'add'] & ['mean'] & ['mean'] & ['mean', 'add'] & ['add']\\ 
 & \textbf{GCN} & ['mean'] & ['mean'] & ['add'] & ['mean'] & ['mean', 'add'] & ['mean', 'add'] & ['mean'] & ['mean', 'add'] & ['mean']\\ 
\cmidrule(lr){2-11}
\multirow{3}{*}{ $\#$8} & \textbf{GAT} & 0.9 & 0.95 & 0.9 & 0.9 & 0.9 & 0.9 & 0.9 & 0.95 & 0.99\\ 
 & \textbf{GIN} & 0.9 & 0.9 & 0.9 & 0.9 & 0.9 & 0.9 & 0.9 & 0.9 & 0.9\\ 
 & \textbf{GCN} & 0.95 & 0.95 & 0.99 & 0.9 & 0.99 & 0.9 & 0.9 & 0.99 & 0.95\\ 
\cmidrule(lr){2-11}
\multirow{3}{*}{ $\#$9} & \textbf{GAT} & 0.005 & 0.0005 & 0.001 & 0.01 & 0.01 & 0.01 & 0.01 & 0.001 & 0.0005\\ 
 & \textbf{GIN} & 0.01 & 0.0001 & 0.001 & 0.01 & 0.01 & 0.01 & 0.01 & 0.001 & 0.0001\\ 
 & \textbf{GCN} & 0.0001 & 0.0001 & 0.01 & 0.01 & 0.01 & 0.001 & 0.01 & 0.001 & 0.005\\ 
\cmidrule(lr){2-11}
\multirow{3}{*}{ $\#$10} & \textbf{GAT} & 2.0 & 0.5 & 1.5 & 2.0 & 2.0 & 2.0 & 2.0 & 1.5 & 0.5\\ 
 & \textbf{GIN} & 2.0 & 2.0 & 1.5 & 2.0 & 1.0 & 2.0 & 2.0 & 1.5 & 0.5\\ 
 & \textbf{GCN} & 1.0 & 1.0 & 1.5 & 2.0 & 0.5 & 1.0 & 2.0 & 0.5 & 1.5\\ 
\cmidrule(lr){2-11}
\multirow{3}{*}{ $\#$11} & \textbf{GAT} & 0.4 & 0.5 & 0.4 & 0.95 & 0.95 & 0.95 & 0.95 & 0.5 & 0.5\\ 
 & \textbf{GIN} & 0.95 & 0.9 & 0.4 & 0.95 & 0.5 & 0.95 & 0.95 & 0.4 & 0.4\\ 
 & \textbf{GCN} & 0.95 & 0.95 & 0.4 & 0.95 & 0.7 & 0.95 & 0.95 & 0.5 & 0.8\\ 
\cmidrule(lr){2-11}
\multirow{3}{*}{ $\#$12} & \textbf{GAT} & 4.0 & 1.0 & 1.0 & 1.0 & 1.0 & 1.0 & 1.0 & 1.0 & 1.0\\ 
 & \textbf{GIN} & - & - & - & - & - & - & - & - & -\\ 
 & \textbf{GCN} & - & - & - & - & - & - & - & - & -\\ 
\cmidrule(lr){2-11}
\multirow{3}{*}{ $\#$13} & \textbf{GAT} & - & - & - & - & - & - & - & - & -\\ 
 & \textbf{GIN} & 0.0 & 0.4 & 0.1 & 0.0 & 0.1 & 0.0 & 0.0 & 0.1 & 0.2\\ 
 & \textbf{GCN} & - & - & - & - & - & - & - & - & -\\ 
\cmidrule(lr){2-11}
\multirow{3}{*}{ $\#$14} & \textbf{GAT} & - & - & - & - & - & - & - & - & -\\ 
 & \textbf{GIN} & True & True & False & True & True & True & True & False & True\\ 
 & \textbf{GCN} & - & - & - & - & - & - & - & - & -\\ 
\bottomrule
\end{tabular}
}
\label{tab:params-topkhard}
\end{table}

\begin{table}[t]
\caption{\textbf{Best configuration for \ours[N].} Early stopping PPO patience ($\#$0), Number of environment steps ($\#$1), Number of PPO epochs ($\#$2), Environment penalty size ($\#$3), RL scheduler factor ($\#$4), Ratio of the critic learning rate ($\#$5), PPO entropy coefficient ($\#$6), PPO MSE coefficient ($\#$7), PPO clip value $\epsilon$ ($\#$8), Conformal error rate $\alpha$ ($\#$9), $d$ ($\#$10), $\lambda$ ($\#$11).}
\centering
{\small
\begin{tabular}{lcccccccccc}
\toprule
\multirow{2}{*}{\textbf{Param}} & \multirow{2}{*}{\textbf{Archi}} & \multicolumn{9}{c}{\textbf{Datasets}} \\ 
\cmidrule(lr){3-11}
& & \textbf{BZR} & \textbf{COX2} & \textbf{DD} & \textbf{ENZYMES} & \textbf{MUTAG} & \textbf{NCI1} & \textbf{NCI109} & \textbf{PROTEINS} & \textbf{PTC}\\ 
\midrule
\multirow{3}{*}{ $\#$0} & \textbf{GIN} & 15 & 5 & 5 & 10 & 10 & 5 & 15 & 5 & 5\\ 
 & \textbf{GCN} & 10 & 5 & 5 & 15 & 5 & 15 & 10 & 15 & 15\\ 
 & \textbf{GAT} & 15 & 5 & 5 & 10 & 15 & 15 & 10 & 10 & 5\\ 
\cmidrule(lr){2-11}
\multirow{3}{*}{ $\#$1} & \textbf{GIN} & 128 & 128 & 128 & 128 & 128 & 128 & 64 & 64 & 64\\ 
 & \textbf{GCN} & 32 & 64 & 64 & 128 & 64 & 32 & 32 & 32 & 128\\ 
 & \textbf{GAT} & 64 & 32 & 32 & 32 & 128 & 64 & 32 & 32 & 64\\ 
\cmidrule(lr){2-11}
\multirow{3}{*}{ $\#$2} & \textbf{GIN} & 10 & 15 & 15 & 5 & 15 & 15 & 10 & 5 & 15\\ 
 & \textbf{GCN} & 10 & 15 & 15 & 5 & 15 & 3 & 10 & 5 & 5\\ 
 & \textbf{GAT} & 10 & 15 & 15 & 10 & 5 & 10 & 10 & 5 & 5\\ 
\cmidrule(lr){2-11}
\multirow{3}{*}{ $\#$3} & \textbf{GIN} & 0.5 & 0.5 & 0.5 & 0.5 & 0.5 & 0.5 & 1.5 & 1.5 & 1.0\\ 
 & \textbf{GCN} & 0.5 & 0.5 & 0.5 & 1.0 & 0.5 & 0.5 & 0.5 & 1.5 & 1.5\\ 
 & \textbf{GAT} & 1.5 & 0.5 & 1.5 & 0.5 & 1.5 & 1.5 & 0.5 & 1.0 & 1.5\\ 
\cmidrule(lr){2-11}
\multirow{3}{*}{ $\#$4} & \textbf{GIN} & 0.99 & 0.95 & 0.95 & 0.99 & 0.9 & 0.95 & 0.9 & 0.9 & 0.95\\ 
 & \textbf{GCN} & 0.99 & 0.95 & 0.95 & 0.99 & 0.95 & 0.99 & 0.99 & 0.9 & 0.9\\ 
 & \textbf{GAT} & 0.9 & 0.95 & 0.99 & 0.99 & 0.9 & 0.9 & 0.99 & 0.95 & 0.9\\ 
\cmidrule(lr){2-11}
\multirow{3}{*}{ $\#$5} & \textbf{GIN} & 3.0 & 2.5 & 2.5 & 1.0 & 3.0 & 2.5 & 2.5 & 1.0 & 3.0\\ 
 & \textbf{GCN} & 2.5 & 1.5 & 1.5 & 2.0 & 1.5 & 3.0 & 2.5 & 2.5 & 1.5\\ 
 & \textbf{GAT} & 2.5 & 2.0 & 2.5 & 2.5 & 1.5 & 2.5 & 2.5 & 1.0 & 1.0\\ 
\cmidrule(lr){2-11}
\multirow{3}{*}{ $\#$6} & \textbf{GIN} & 0.001 & 0.0001 & 0.0001 & 0.01 & 0.001 & 0.0001 & 0.0001 & 0.001 & 0.0001\\ 
 & \textbf{GCN} & 0.0001 & 0.01 & 0.01 & 0.0001 & 0.01 & 0.01 & 0.0001 & 0.01 & 0.001\\ 
 & \textbf{GAT} & 0.0001 & 0.001 & 0.001 & 0.0001 & 0.001 & 0.0001 & 0.0001 & 0.001 & 0.001\\ 
\cmidrule(lr){2-11}
\multirow{3}{*}{ $\#$7} & \textbf{GIN} & 0.5 & 5.0 & 5.0 & 2.0 & 1.0 & 5.0 & 2.0 & 3.0 & 0.5\\ 
 & \textbf{GCN} & 1.0 & 0.5 & 0.5 & 0.1 & 0.5 & 1.0 & 1.0 & 5.0 & 3.0\\ 
 & \textbf{GAT} & 2.0 & 1.0 & 3.0 & 1.0 & 3.0 & 2.0 & 1.0 & 0.1 & 3.0\\ 
\cmidrule(lr){2-11}
\multirow{3}{*}{ $\#$8} & \textbf{GIN} & 0.2 & 0.4 & 0.4 & 0.1 & 0.2 & 0.4 & 0.2 & 0.1 & 0.4\\ 
 & \textbf{GCN} & 0.4 & 0.3 & 0.3 & 0.3 & 0.3 & 0.1 & 0.4 & 0.3 & 0.1\\ 
 & \textbf{GAT} & 0.2 & 0.2 & 0.4 & 0.4 & 0.1 & 0.2 & 0.4 & 0.4 & 0.1\\ 
\cmidrule(lr){2-11}
\multirow{3}{*}{ $\#$9} & \textbf{GIN} & 0.2 & 0.05 & 0.05 & 0.1 & 0.2 & 0.05 & 0.2 & 0.05 & 0.05\\ 
 & \textbf{GCN} & 0.05 & 0.2 & 0.2 & 0.05 & 0.2 & 0.1 & 0.05 & 0.2 & 0.05\\ 
 & \textbf{GAT} & 0.2 & 0.2 & 0.2 & 0.05 & 0.05 & 0.2 & 0.05 & 0.05 & 0.05\\ 
\cmidrule(lr){2-11}
\multirow{3}{*}{ $\#$10} & \textbf{GIN} & 0.2 & 0.3 & 0.3 & 0.7 & 0.7 & 0.3 & 0.4 & 0.2 & 0.5\\ 
 & \textbf{GCN} & 0.9 & 0.8 & 0.8 & 0.6 & 0.8 & 0.4 & 0.9 & 0.5 & 0.2\\ 
 & \textbf{GAT} & 0.4 & 0.6 & 0.6 & 0.9 & 0.2 & 0.4 & 0.9 & 0.6 & 0.2\\ 
\cmidrule(lr){2-11}
\multirow{3}{*}{ $\#$11} & \textbf{GIN} & 0.2 & 1.0 & 1.0 & 0.9 & 0.1 & 1.0 & 0.4 & 0.0 & 0.4\\ 
 & \textbf{GCN} & 0.2 & 0.7 & 0.7 & 1.0 & 0.7 & 0.8 & 0.2 & 0.8 & 0.7\\ 
 & \textbf{GAT} & 0.4 & 0.3 & 0.8 & 0.2 & 0.7 & 0.4 & 0.2 & 0.5 & 0.0\\ 
\bottomrule
\end{tabular}
}
\label{tab:params-cores-node}
\end{table}

\begin{table}[t]
\caption{\textbf{Best configuration for \ours[E].} Early stopping PPO patience ($\#$0), Number of environment steps ($\#$1), Number of PPO epochs ($\#$2), Environment penalty size ($\#$3), RL scheduler factor ($\#$4), Ratio of the critic learning rate ($\#$5), PPO entropy coefficient ($\#$6), PPO MSE coefficient ($\#$7), PPO clip value $\epsilon$ ($\#$8), Conformal error rate $\alpha$ ($\#$9), $d$ ($\#$10), $\lambda$ ($\#$11).}
\centering
{\small
\begin{tabular}{lcccccccccc}
\toprule
\multirow{2}{*}{\textbf{Param}} & \multirow{2}{*}{\textbf{Archi}} & \multicolumn{9}{c}{\textbf{Datasets}} \\ 
\cmidrule(lr){3-11}
& & \textbf{BZR} & \textbf{COX2} & \textbf{DD} & \textbf{ENZYMES} & \textbf{MUTAG} & \textbf{NCI1} & \textbf{NCI109} & \textbf{PROTEINS} & \textbf{PTC}\\ 
\midrule
\multirow{3}{*}{ $\#$0} & \textbf{GIN} & 15 & 5 & 5 & 10 & 10 & 5 & 15 & 5 & 5\\ 
 & \textbf{GCN} & 10 & 5 & 5 & 15 & 5 & 15 & 10 & 15 & 15\\ 
 & \textbf{GAT} & 15 & 5 & 5 & 10 & 15 & 15 & 10 & 10 & 5\\ 
\cmidrule(lr){2-11}
\multirow{3}{*}{ $\#$1} & \textbf{GIN} & 128 & 128 & 128 & 128 & 128 & 128 & 64 & 64 & 64\\ 
 & \textbf{GCN} & 32 & 64 & 64 & 128 & 64 & 32 & 32 & 32 & 128\\ 
 & \textbf{GAT} & 64 & 32 & 32 & 32 & 128 & 64 & 32 & 32 & 64\\ 
\cmidrule(lr){2-11}
\multirow{3}{*}{ $\#$2} & \textbf{GIN} & 10 & 15 & 15 & 5 & 15 & 15 & 10 & 5 & 15\\ 
 & \textbf{GCN} & 10 & 15 & 15 & 5 & 15 & 3 & 10 & 5 & 5\\ 
 & \textbf{GAT} & 10 & 15 & 15 & 10 & 5 & 10 & 10 & 5 & 5\\ 
\cmidrule(lr){2-11}
\multirow{3}{*}{ $\#$3} & \textbf{GIN} & 0.5 & 0.5 & 0.5 & 0.5 & 0.5 & 0.5 & 1.5 & 1.5 & 1.0\\ 
 & \textbf{GCN} & 0.5 & 0.5 & 0.5 & 1.0 & 0.5 & 0.5 & 0.5 & 1.5 & 1.5\\ 
 & \textbf{GAT} & 1.5 & 0.5 & 1.5 & 0.5 & 1.5 & 1.5 & 0.5 & 1.0 & 1.5\\ 
\cmidrule(lr){2-11}
\multirow{3}{*}{ $\#$4} & \textbf{GIN} & 0.99 & 0.95 & 0.95 & 0.99 & 0.9 & 0.95 & 0.9 & 0.9 & 0.95\\ 
 & \textbf{GCN} & 0.99 & 0.95 & 0.95 & 0.99 & 0.95 & 0.99 & 0.99 & 0.9 & 0.9\\ 
 & \textbf{GAT} & 0.9 & 0.95 & 0.99 & 0.99 & 0.9 & 0.9 & 0.99 & 0.95 & 0.9\\ 
\cmidrule(lr){2-11}
\multirow{3}{*}{ $\#$5} & \textbf{GIN} & 3.0 & 2.5 & 2.5 & 1.0 & 3.0 & 2.5 & 2.5 & 1.0 & 3.0\\ 
 & \textbf{GCN} & 2.5 & 1.5 & 1.5 & 2.0 & 1.5 & 3.0 & 2.5 & 2.5 & 1.5\\ 
 & \textbf{GAT} & 2.5 & 2.0 & 2.5 & 2.5 & 1.5 & 2.5 & 2.5 & 1.0 & 1.0\\ 
\cmidrule(lr){2-11}
\multirow{3}{*}{ $\#$6} & \textbf{GIN} & 0.001 & 0.0001 & 0.0001 & 0.01 & 0.001 & 0.0001 & 0.0001 & 0.001 & 0.0001\\ 
 & \textbf{GCN} & 0.0001 & 0.01 & 0.01 & 0.0001 & 0.01 & 0.01 & 0.0001 & 0.01 & 0.001\\ 
 & \textbf{GAT} & 0.0001 & 0.001 & 0.001 & 0.0001 & 0.001 & 0.0001 & 0.0001 & 0.001 & 0.001\\ 
\cmidrule(lr){2-11}
\multirow{3}{*}{ $\#$7} & \textbf{GIN} & 0.5 & 5.0 & 5.0 & 2.0 & 1.0 & 5.0 & 2.0 & 3.0 & 0.5\\ 
 & \textbf{GCN} & 1.0 & 0.5 & 0.5 & 0.1 & 0.5 & 1.0 & 1.0 & 5.0 & 3.0\\ 
 & \textbf{GAT} & 2.0 & 1.0 & 3.0 & 1.0 & 3.0 & 2.0 & 1.0 & 0.1 & 3.0\\ 
\cmidrule(lr){2-11}
\multirow{3}{*}{ $\#$8} & \textbf{GIN} & 0.2 & 0.4 & 0.4 & 0.1 & 0.2 & 0.4 & 0.2 & 0.1 & 0.4\\ 
 & \textbf{GCN} & 0.4 & 0.3 & 0.3 & 0.3 & 0.3 & 0.1 & 0.4 & 0.3 & 0.1\\ 
 & \textbf{GAT} & 0.2 & 0.2 & 0.4 & 0.4 & 0.1 & 0.2 & 0.4 & 0.4 & 0.1\\ 
\cmidrule(lr){2-11}
\multirow{3}{*}{ $\#$9} & \textbf{GIN} & 0.2 & 0.05 & 0.05 & 0.1 & 0.2 & 0.05 & 0.2 & 0.05 & 0.05\\ 
 & \textbf{GCN} & 0.05 & 0.2 & 0.2 & 0.05 & 0.2 & 0.1 & 0.05 & 0.2 & 0.05\\ 
 & \textbf{GAT} & 0.2 & 0.2 & 0.2 & 0.05 & 0.05 & 0.2 & 0.05 & 0.05 & 0.05\\ 
\cmidrule(lr){2-11}
\multirow{3}{*}{ $\#$10} & \textbf{GIN} & 0.2 & 0.3 & 0.3 & 0.7 & 0.7 & 0.3 & 0.4 & 0.2 & 0.5\\ 
 & \textbf{GCN} & 0.9 & 0.8 & 0.8 & 0.6 & 0.8 & 0.4 & 0.9 & 0.5 & 0.2\\ 
 & \textbf{GAT} & 0.4 & 0.6 & 0.6 & 0.9 & 0.2 & 0.4 & 0.9 & 0.6 & 0.2\\ 
\cmidrule(lr){2-11}
\multirow{3}{*}{ $\#$11} & \textbf{GIN} & 0.2 & 1.0 & 1.0 & 0.9 & 0.1 & 1.0 & 0.4 & 0.0 & 0.4\\ 
 & \textbf{GCN} & 0.2 & 0.7 & 0.7 & 1.0 & 0.7 & 0.8 & 0.2 & 0.8 & 0.7\\ 
 & \textbf{GAT} & 0.4 & 0.3 & 0.8 & 0.2 & 0.7 & 0.4 & 0.2 & 0.5 & 0.0\\ 
\bottomrule
\end{tabular}
}
\label{tab:params-cores-edge}
\end{table}

\clearpage
\newpage

\section{Extra results impact of hyperparameters}
\label{app:ablation-hyper}

\begin{figure}[tbp]
  \centering
  \includegraphics[width=0.215\textwidth]{figures/legends_model_legend.pdf}
  \includegraphics[width=0.5\textwidth]{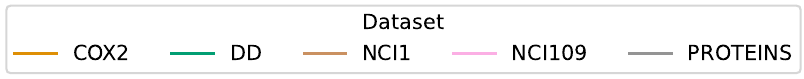}

  \begin{minipage}{0.49\textwidth}
    \centering
    \begin{subfigure}[b]{1.0\textwidth}
        \begin{subfigure}[b]{0.49\textwidth}
        \includegraphics[width=\textwidth]{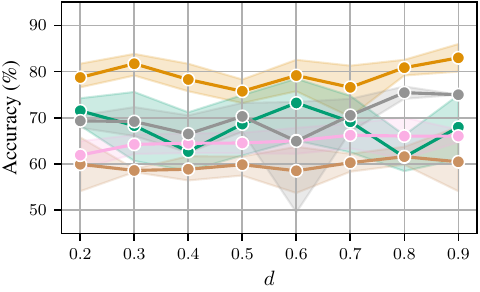}
        \end{subfigure}
        \begin{subfigure}[b]{0.49\textwidth}
        \includegraphics[width=\textwidth]{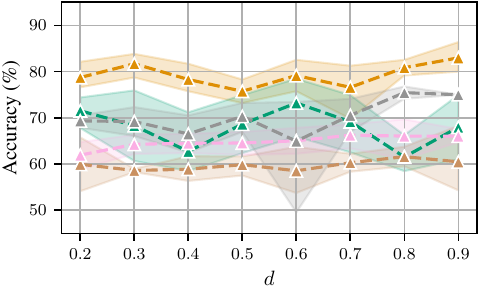}
        \end{subfigure}
         \caption{$d$ versus accuracy}
    \label{fig:d_perf}
    \end{subfigure}
     \begin{subfigure}[b]{1.0\textwidth}
        \begin{subfigure}[b]{0.49\textwidth}
        \includegraphics[width=\textwidth]{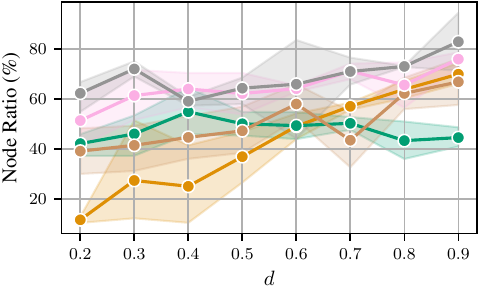}
        \end{subfigure}
        \begin{subfigure}[b]{0.49\textwidth}
        \includegraphics[width=\textwidth]{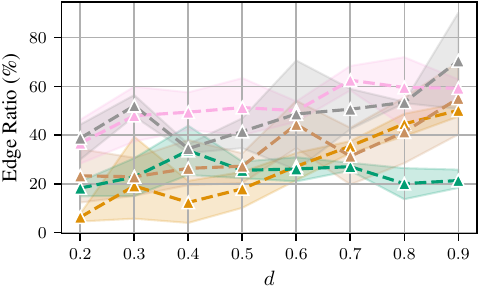}
        \end{subfigure}
        \caption{$d$ versus sparsity}
    \label{fig:d_spar}
    \end{subfigure}
   \begin{subfigure}[b]{1.0\textwidth}
    \begin{subfigure}[b]{0.49\textwidth}
        \includegraphics[width=\textwidth]{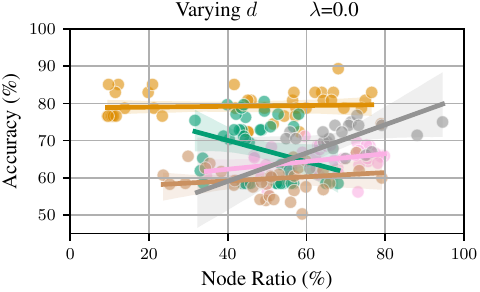}
    \end{subfigure}
    \begin{subfigure}[b]{0.49\textwidth}
        \includegraphics[width=\textwidth]{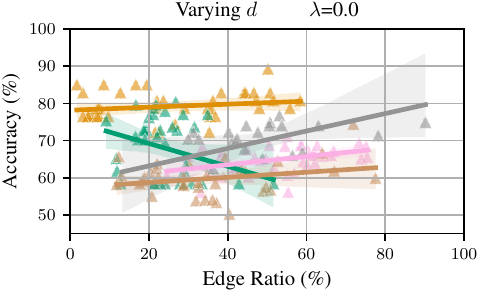}
    \end{subfigure}
    \caption{Sparsity versus accuracy}
    \end{subfigure}
    \caption{Ablation study on the maximum desired ratio} $d$ for 5 different runs for the \gin\ architecture.
    \label{figapp:ablation_d}
  \end{minipage}
  \hfill
  \begin{minipage}{0.49\textwidth}
    \centering
    \begin{subfigure}[b]{1.0\textwidth}
        \begin{subfigure}[b]{0.49\textwidth}
        \includegraphics[width=\textwidth]{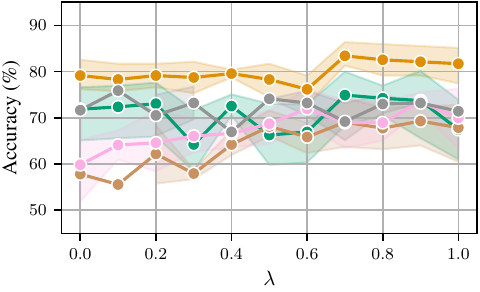}
      \end{subfigure}
      \begin{subfigure}[b]{0.49\textwidth}
        \includegraphics[width=\textwidth]{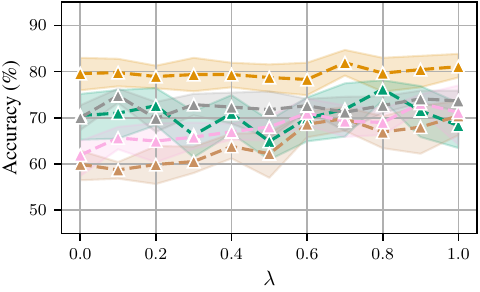}
      \end{subfigure}
      \caption{$\lambda$ versus accuracy}
  \end{subfigure}
      \begin{subfigure}[b]{1.0\textwidth}
        \begin{subfigure}[b]{0.49\textwidth}
        \includegraphics[width=\textwidth]{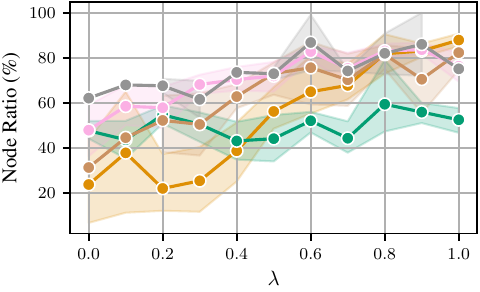}
      \end{subfigure}
      \begin{subfigure}[b]{0.49\textwidth}
        \includegraphics[width=\textwidth]{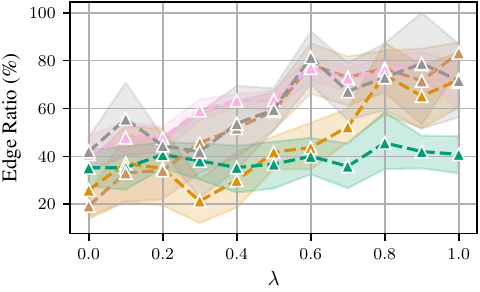}
      \end{subfigure}
      \caption{$\lambda$ versus sparsity}
  \end{subfigure}
      \begin{subfigure}[b]{1.0\textwidth}
        \begin{subfigure}[b]{0.49\textwidth}
            \includegraphics[width=\textwidth]{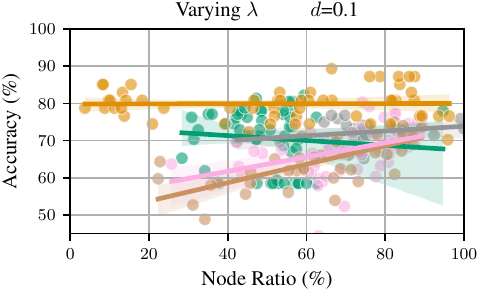}
        \end{subfigure}
        \begin{subfigure}[b]{0.49\textwidth}
            \includegraphics[width=\textwidth]{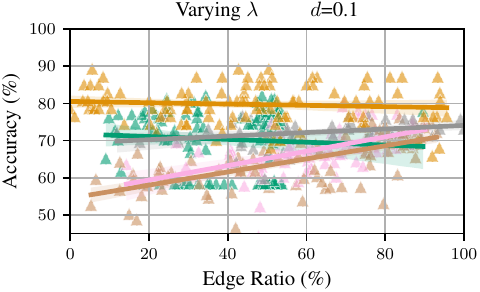}
        \end{subfigure}
        \caption{Sparsity versus accuracy}
    \end{subfigure}
  
    \caption{Ablation study on $\lambda$ for 5 different runs for the \gin\ architecture.}
    \label{figapp:ablation_lambda}
  \end{minipage}
\end{figure}

In this section, we present an ablation study on the impact of two key hyperparameters of \ours, namely $d$ and $\lambda$, for the datasets not included in the main manuscript: \cox, \dd, \ncione, \nciten, and \proteins. 
\cref{figapp:ablation_d} illustrates the results of the ablation study for parameter $d$, while the results for parameter $\lambda$ are shown in \cref{figapp:ablation_lambda}.

The findings from this ablation study are consistent with those reported in the main manuscript. 
In the top row figures, we observe that the accuracy remains relatively stable when varying either $d$ or $\lambda$, improving only as the values increase for certain datasets, such as \ncione or \nciten.
In the middle row figures, we find a positive correlation between both parameters and the node/edge ratio for most of the datasets, and no correlation for others. Remarkably, the correlation is never negative.
Finally, the bottom row figures reveal a positive correlation between the node/edge ratio and accuracy for the majority of the datasets. 
This pattern holds except for the \dd\ dataset, which present a negative correlation. Interestingly this dataset presents the most significant disparities in terms of the number of nodes and edges compared to the other datasets (see Table \ref{tab:dataset-stats}).

	\clearpage
	\newpage

\section{Extra comparison results}
\label{app:comparison}

Here, we analyze the results of how switching the base model affects the performance. \cref{tabapp:comparison-gcn}  and \cref{tabapp:comparison-gat} summarizes the results for two base architectures \gcn\ and \gat. It can be observed that even when the base architecture changes our models \ours[N] and \ours[E] achieve competitive accuracy compared to the respective baseline and achieve top rankings in terms of node/edge ratio.

\begin{table}[t]
\caption{\textbf{Model comparison results with \gat\ architecture.} We show the mean over 5 independent runs and the standard deviation as the subindex. All metrics are shown in percentage. The last rows include the average ranking of the model across datasets.}\centering
{\small
\begin{tabular}{lcccccc}
\toprule
\multirow{2}{*}{\textbf{Dataset}} & \multirow{2}{*}{\textbf{Metric}} & \multicolumn{5}{c}{\textbf{Models}} \\ 
\cmidrule(lr){3-7}
& & \textbf{GNN} & \textbf{SAGPool} & \textbf{GPool} & \textbf{CORES$_N$} & \textbf{CORES$_E$}\\ 
\midrule
\multirow{3}{*}{BZR} & \textbf{Accuracy} & 80.98$_{4.36}$ & 83.90$_{3.27}$ & 79.51$_{6.12}$ & 83.90$_{4.43}$ & 81.46$_{4.08}$\\ 
 & \textbf{Node Ratio} & - & - & 41.11$_{0.11}$ & 37.99$_{10.27}$ & 100.00$_{0.00}$\\ 
 & \textbf{Edge Ratio} & - & - & 20.93$_{0.70}$ & 29.92$_{8.83}$ & 57.26$_{21.50}$\\ 
\cmidrule(lr){2-7}
\multirow{3}{*}{COX2} & \textbf{Accuracy} & 82.13$_{4.15}$ & 80.00$_{4.41}$ & 77.87$_{3.87}$ & 81.70$_{4.41}$ & 80.85$_{2.61}$\\ 
 & \textbf{Node Ratio} & - & - & 50.62$_{0.09}$ & 77.91$_{2.29}$ & 100.00$_{0.00}$\\ 
 & \textbf{Edge Ratio} & - & - & 20.75$_{0.51}$ & 62.62$_{3.24}$ & 66.18$_{0.38}$\\ 
\cmidrule(lr){2-7}
\multirow{3}{*}{DD} & \textbf{Accuracy} & 76.10$_{3.02}$ & 75.76$_{3.72}$ & 77.80$_{1.84}$ & 75.42$_{4.19}$ & 76.44$_{2.19}$\\ 
 & \textbf{Node Ratio} & - & - & 40.24$_{0.03}$ & 75.16$_{1.39}$ & 100.00$_{0.00}$\\ 
 & \textbf{Edge Ratio} & - & - & 15.63$_{0.29}$ & 56.42$_{2.16}$ & 58.62$_{2.51}$\\ 
\cmidrule(lr){2-7}
\multirow{3}{*}{ENZYMES} & \textbf{Accuracy} & 71.33$_{10.63}$ & 45.00$_{7.64}$ & 63.93$_{6.35}$ & 69.84$_{4.43}$ & 67.54$_{12.23}$\\ 
 & \textbf{Node Ratio} & - & - & 96.67$_{0.14}$ & 80.26$_{1.03}$ & 100.00$_{0.00}$\\ 
 & \textbf{Edge Ratio} & - & - & 93.27$_{0.23}$ & 65.92$_{1.80}$ & 61.60$_{9.06}$\\ 
\cmidrule(lr){2-7}
\multirow{3}{*}{MUTAG} & \textbf{Accuracy} & 81.43$_{6.39}$ & 88.57$_{3.91}$ & 78.57$_{7.14}$ & 80.00$_{5.98}$ & 81.43$_{6.02}$\\ 
 & \textbf{Node Ratio} & - & - & 97.81$_{0.36}$ & 75.72$_{4.04}$ & 100.00$_{0.00}$\\ 
 & \textbf{Edge Ratio} & - & - & 95.65$_{2.50}$ & 58.74$_{6.05}$ & 55.25$_{14.77}$\\ 
\cmidrule(lr){2-7}
\multirow{3}{*}{NCI1} & \textbf{Accuracy} & 79.42$_{1.34}$ & 68.93$_{1.37}$ & 72.43$_{2.11}$ & 73.43$_{2.67}$ & 75.38$_{1.22}$\\ 
 & \textbf{Node Ratio} & - & - & 96.89$_{0.08}$ & 63.49$_{3.58}$ & 100.00$_{0.00}$\\ 
 & \textbf{Edge Ratio} & - & - & 94.07$_{0.34}$ & 53.33$_{2.81}$ & 88.68$_{6.67}$\\ 
\cmidrule(lr){2-7}
\multirow{3}{*}{NCI109} & \textbf{Accuracy} & 76.90$_{1.34}$ & 67.46$_{4.31}$ & 67.22$_{3.25}$ & 74.72$_{2.25}$ & 72.78$_{2.29}$\\ 
 & \textbf{Node Ratio} & - & - & 96.79$_{0.04}$ & 71.55$_{11.30}$ & 100.00$_{0.00}$\\ 
 & \textbf{Edge Ratio} & - & - & 94.70$_{0.98}$ & 62.37$_{12.96}$ & 61.37$_{13.13}$\\ 
\cmidrule(lr){2-7}
\multirow{3}{*}{PROTEINS} & \textbf{Accuracy} & 75.36$_{3.00}$ & 72.68$_{3.26}$ & 73.21$_{3.22}$ & 73.39$_{3.86}$ & 73.57$_{1.62}$\\ 
 & \textbf{Node Ratio} & - & - & 51.15$_{0.17}$ & 67.94$_{2.97}$ & 100.00$_{0.00}$\\ 
 & \textbf{Edge Ratio} & - & - & 28.39$_{0.42}$ & 46.56$_{4.27}$ & 52.82$_{1.86}$\\ 
\cmidrule(lr){2-7}
\multirow{3}{*}{PTC} & \textbf{Accuracy} & 64.00$_{4.33}$ & 65.00$_{6.39}$ & 62.44$_{4.12}$ & 62.96$_{4.17}$ & 59.43$_{4.69}$\\ 
 & \textbf{Node Ratio} & - & - & 52.46$_{0.41}$ & 45.61$_{6.00}$ & 100.00$_{0.00}$\\ 
 & \textbf{Edge Ratio} & - & - & 36.58$_{3.53}$ & 21.24$_{6.53}$ & 76.36$_{14.75}$\\ 
\bottomrule
\multirow{3}{*}{\textbf{Avg. Rank}} & \textbf{Accuracy} & 1.89 & 3.33 & 3.89 & 3.00 & 2.89\\ 
\cmidrule(lr){2-7}
 & \textbf{Node Ratio} & - & - & 1.67 & 1.33 & - \\ 
\cmidrule(lr){2-7}
 & \textbf{Edge Ratio} & - & - & 2.33 & 1.56 & 2.11\\ 
\bottomrule
\end{tabular}
}
\label{tabapp:comparison-gat}
\end{table}

\begin{table}[t]
\caption{\textbf{Model comparison results with \gcn\ architecture} We show the mean over 5 independent runs and the standard deviation as the subindex. All metrics are shown in percentage. The last rows include the average ranking of the model across datasets.}
\centering
{\small
\begin{tabular}{lcccccc}
\toprule
\multirow{2}{*}{\textbf{Dataset}} & \multirow{2}{*}{\textbf{Metric}} & \multicolumn{5}{c}{\textbf{Models}} \\ 
\cmidrule(lr){3-7}
& & \textbf{GNN} & \textbf{SAGPool} & \textbf{GPool} & \textbf{CORES$_N$} & \textbf{CORES$_E$}\\ 
\midrule
\multirow{3}{*}{BZR} & \textbf{Accuracy} & 86.83$_{4.08}$ & 83.41$_{4.69}$ & 84.39$_{4.08}$ & 82.44$_{4.69}$ & 80.98$_{5.29}$\\ 
 & \textbf{Node Ratio} & - & - & 96.46$_{0.19}$ & 76.98$_{6.06}$ & 100.00$_{0.00}$\\ 
 & \textbf{Edge Ratio} & - & - & 92.87$_{0.58}$ & 61.17$_{11.31}$ & 46.16$_{5.68}$\\ 
\cmidrule(lr){2-7}
\multirow{3}{*}{COX2} & \textbf{Accuracy} & 79.15$_{4.85}$ & 80.43$_{3.81}$ & 80.00$_{1.90}$ & 80.85$_{3.01}$ & 81.70$_{4.90}$\\ 
 & \textbf{Node Ratio} & - & - & 96.14$_{0.18}$ & 72.49$_{4.24}$ & 100.00$_{0.00}$\\ 
 & \textbf{Edge Ratio} & - & - & 92.67$_{0.71}$ & 53.19$_{8.07}$ & 31.27$_{23.60}$\\ 
\cmidrule(lr){2-7}
\multirow{3}{*}{DD} & \textbf{Accuracy} & 78.47$_{3.47}$ & 75.93$_{4.39}$ & 76.10$_{5.06}$ & 79.32$_{2.58}$ & 75.76$_{4.05}$\\ 
 & \textbf{Node Ratio} & - & - & 40.23$_{0.03}$ & 56.44$_{2.50}$ & 100.00$_{0.00}$\\ 
 & \textbf{Edge Ratio} & - & - & 16.48$_{0.13}$ & 32.01$_{2.89}$ & 51.39$_{1.90}$\\ 
\cmidrule(lr){2-7}
\multirow{3}{*}{ENZYMES} & \textbf{Accuracy} & 66.00$_{5.35}$ & 52.00$_{5.70}$ & 64.59$_{4.86}$ & 58.31$_{8.47}$ & 57.02$_{11.62}$\\ 
 & \textbf{Node Ratio} & - & - & 96.67$_{0.14}$ & 77.17$_{7.28}$ & 100.00$_{0.00}$\\ 
 & \textbf{Edge Ratio} & - & - & 93.39$_{0.35}$ & 62.33$_{9.26}$ & 74.20$_{35.99}$\\ 
\cmidrule(lr){2-7}
\multirow{3}{*}{MUTAG} & \textbf{Accuracy} & 81.43$_{3.91}$ & 80.00$_{3.19}$ & 78.57$_{5.05}$ & 81.43$_{10.83}$ & 87.14$_{6.56}$\\ 
 & \textbf{Node Ratio} & - & - & 73.13$_{0.55}$ & 79.94$_{2.79}$ & 100.00$_{0.00}$\\ 
 & \textbf{Edge Ratio} & - & - & 68.22$_{2.47}$ & 63.52$_{4.23}$ & 32.66$_{8.79}$\\ 
\cmidrule(lr){2-7}
\multirow{3}{*}{NCI1} & \textbf{Accuracy} & 77.33$_{2.09}$ & 73.58$_{1.64}$ & 72.14$_{2.64}$ &  66.45$_{7.71}$ & 65.94$_{6.31}$ \\ 
 & \textbf{Node Ratio} & - & - & 96.89$_{0.08}$ & 76.47$_{17.88}$ & 100.00$_{0.00}$\\ 
 & \textbf{Edge Ratio} & - & - & 96.06$_{0.20}$ & 64.02$_{18.84}$ & 37.98$_{14.34}$\\ 
\cmidrule(lr){2-7}
\multirow{3}{*}{NCI109} & \textbf{Accuracy} & 79.32$_{1.57}$ & 71.19$_{2.03}$ & 71.53$_{2.12}$ & 76.08$_{2.71}$ & 77.14$_{2.02}$\\ 
 & \textbf{Node Ratio} & - & - & 96.79$_{0.04}$ & 78.78$_{4.15}$ & 100.00$_{0.00}$\\ 
 & \textbf{Edge Ratio} & - & - & 95.82$_{0.11}$ & 67.08$_{4.42}$ & 92.02$_{1.93}$\\ 
\cmidrule(lr){2-7}
\multirow{3}{*}{PROTEINS} & \textbf{Accuracy} & 72.86$_{1.02}$ & 72.86$_{3.13}$ & 75.18$_{2.75}$ & 73.21$_{2.45}$ & 75.89$_{1.94}$\\ 
 & \textbf{Node Ratio} & - & - & 51.15$_{0.17}$ & 74.66$_{6.10}$ & 100.00$_{0.00}$\\ 
 & \textbf{Edge Ratio} & - & - & 28.15$_{0.67}$ & 56.12$_{9.77}$ & 64.73$_{0.98}$\\ 
\cmidrule(lr){2-7}
\multirow{3}{*}{PTC} & \textbf{Accuracy} & 57.78$_{7.71}$ & 64.57$_{3.83}$ & 63.43$_{1.28}$ & 62.86$_{2.86}$ & 63.43$_{3.13}$\\ 
 & \textbf{Node Ratio} & - & - & 83.80$_{0.74}$ & 64.10$_{4.19}$ & 100.00$_{0.00}$\\ 
 & \textbf{Edge Ratio} & - & - & 78.21$_{3.76}$ & 40.21$_{6.93}$ & 44.47$_{5.29}$\\ 
\bottomrule
\multirow{3}{*}{\textbf{Avg. Rank}} & \textbf{Accuracy} & 1.89 & 3.33 & 3.89 & 3.00 & 2.89\\ 
\cmidrule(lr){2-7}
 & \textbf{Node Ratio} & - & - & 1.67 & 1.33 & - \\ 
\cmidrule(lr){2-7}
 & \textbf{Edge Ratio} & - & - &  2.56 & 1.67 & 1.78\\ 
\bottomrule
\end{tabular}
}
\label{tabapp:comparison-gcn}
\end{table}

\clearpage
\newpage

\section{Time complexity}
\label{app:time-complexity}

We analyze the training and inference times of all models on each dataset. We run each model on every dataset for a single epoch, utilizing a batch size of one. To ensure statistical significance and reliability in our measurements, we repeat each experiment ten times. This repetition enables us to calculate the average training and inference times accurately. Such a meticulous approach guarantees robust and precise performance assessment for our models across the diverse set of datasets under examination.

\cref{tabapp:time-training} and \cref{tabapp:time-inference} show the complete results for the training and inferences times, respectively. 
If we focus on \cref{tabapp:time-training}, we observe the models that only rely on a single forward pass are two orders of magnitude faster than the models that rely on a reinforcement learning-based approach. Still, we observe that \ours\ achieves comparable speed as \sugar, and sometimes it is faster, e.g., with \ptc\ or \mutag.
In terms of inference time, in \cref{tabapp:time-inference}, we can observe that even though \ours\ is still one order of magnitude slower than the \textit{Full Models}, it is considerably faster than \sugar.

\begin{table}[t]
\caption{\textbf{Training time comparison.} We show the average training time in seconds along with the standard deviation of each model on each dataset for one epoch with a batch size of one. We use the \gin\ GNN architecture.}
\centering
{\small
\begin{tabular}{lrr|rrrr}
\toprule
\textbf{Dataset} & \multicolumn{2}{c}{\textbf{Full Models}} & \multicolumn{4}{c}{\textbf{Sparse Models}} \\ 
\cmidrule(lr){2 - 3} \cmidrule(lr){4 - 7}
& \textbf{\vanilla} & \textbf{\diffpool} & \textbf{\sugar} & \textbf{\topkhard} & \textbf{\ours[N]}   & \textbf{\ours[E]} \\ 
\midrule
\bzr & 2.64 $_{0.18}$ & 4.23 $_{0.12}$ & - & 3.44 $_{0.27}$ & 186.96 $_{61.70}$ & 121.94 $_{2.00}$ \\ 
\cox & 3.07 $_{0.15}$ & 4.83 $_{0.17}$ & - & 10.00 $_{2.70}$ & 411.93 $_{85.97}$ & 203.93 $_{1.86}$ \\ 
\dd & 6.13 $_{0.13}$ & - & - & 33.31 $_{4.90}$ & 1848.21 $_{1.12}$ & 1857.59 $_{1.13}$ \\ 
\enzymes & 2.40 $_{0.04}$ & 6.26 $_{0.31}$ & 132.86 $_{3.77}$ & 6.08 $_{0.30}$ & 233.78 $_{6.65}$ & 287.95 $_{2.20}$ \\ 
\mutag & 1.09 $_{0.03}$ & 1.63 $_{0.04}$ & 47.82 $_{3.45}$ & 1.71 $_{0.19}$ & 24.97 $_{1.55}$ & 22.94 $_{0.30}$ \\ 
\ncione & 15.44 $_{0.19}$ & 41.74 $_{3.03}$ & 1143.85 $_{45.27}$ & 48.05 $_{8.12}$ & 1596.95 $_{107.83}$ & 1826.41 $_{51.06}$ \\ 
\nciten & 16.68 $_{1.15}$ & 43.73 $_{0.44}$ & 1195.10 $_{47.42}$ & 43.46 $_{3.09}$ & 1533.59 $_{15.51}$ & 1756.84 $_{37.12}$ \\ 
\proteins & 5.80 $_{0.23}$ & 11.79 $_{1.19}$ & 511.21 $_{22.63}$ & 13.58 $_{2.58}$ & 484.97 $_{14.74}$ & 505.89 $_{2.70}$ \\ 
\ptc & 1.68 $_{0.16}$ & 3.00 $_{0.07}$ & 95.78 $_{1.47}$ & 3.28 $_{0.18}$ & 64.60 $_{1.23}$ & 67.19 $_{0.32}$ \\ 
\bottomrule
\end{tabular}
}
\label{tabapp:time-training}
\end{table}

\begin{table}[ht]
\caption{\textbf{Inference time comparison.} We show the average inference time in seconds along with the standard deviation of each model on each dataset. We use the \gin\ GNN architecture. }
\centering
{\small
\begin{tabular}{lrr|rrrr}
\toprule
\textbf{Dataset} & \multicolumn{2}{c}{\textbf{Full Models}} & \multicolumn{4}{c}{\textbf{Sparse Models}} \\ 
\cmidrule(lr){2 - 3} \cmidrule(lr){4 - 7}
& \textbf{\vanilla} & \textbf{\diffpool} & \textbf{\sugar} & \textbf{\topkhard} & \textbf{\ours[N]}   & \textbf{\ours[E]} \\ 
\midrule
\bzr & 0.0025 $_{0.0001}$ & 0.1447 $_{0.0048}$ & - & 0.0045 $_{0.0003}$ & 0.0540 $_{0.0257}$ & 0.0425 $_{0.0011}$ \\ 
\cox & 0.0026 $_{0.0001}$ & 0.1550 $_{0.0093}$ & - & 0.0062 $_{0.0008}$ & 0.1031 $_{0.0349}$ & 0.0526 $_{0.0014}$ \\ 
\dd & 0.0028 $_{0.0004}$ & - & - & 0.0085 $_{0.0006}$ & 0.2262 $_{0.0842}$ & 0.1206 $_{0.0016}$ \\ 
\enzymes & 0.0024 $_{0.0000}$ & 0.1992 $_{0.0080}$ & 0.1399 $_{0.0094}$ & 0.0063 $_{0.0007}$ & 0.0496 $_{0.0014}$ & 0.0625 $_{0.0019}$ \\ 
\mutag & 0.0029 $_{0.0001}$ & 0.0596 $_{0.0034}$ & 0.0135 $_{0.0036}$ & 0.0065 $_{0.0017}$ & 0.0448 $_{0.0056}$ & 0.0415 $_{0.0012}$ \\ 
\ncione & 0.0022 $_{0.0001}$ & 1.3756 $_{0.1246}$ & 1.0187 $_{0.1040}$ & 0.0058 $_{0.0004}$ & 0.0530 $_{0.0026}$ & 0.1301 $_{0.0602}$ \\ 
\nciten & 0.0023 $_{0.0001}$ & 1.3685 $_{0.0988}$ & 0.9477 $_{0.0132}$ & 0.0055 $_{0.0002}$ & 0.0500 $_{0.0006}$ & 0.0563 $_{0.0009}$ \\ 
\proteins & 0.0025 $_{0.0001}$ & 0.3835 $_{0.0619}$ & 3.9375 $_{0.0347}$ & 0.0058 $_{0.0007}$ & 0.0437 $_{0.0016}$ & 0.0439 $_{0.0008}$ \\ 
\ptc & 0.0028 $_{0.0001}$ & 0.1076 $_{0.0050}$ & 0.0398 $_{0.0099}$ & 0.0058 $_{0.0005}$ & 0.0395 $_{0.0013}$ & 0.0418 $_{0.0005}$ \\ 
\bottomrule
\end{tabular}
}
\label{tabapp:time-inference}
\end{table}

\end{document}